\documentclass{article}

\usepackage{graphicx} % more modern
\usepackage{subfigure, enumitem}
\usepackage{pgf}
\usepackage{tikz}
\usepackage{pgfplots}
\pgfplotsset{}
\usepackage{booktabs, multicol, multirow}
\usepackage{amsmath}
\makeatletter
\g@addto@macro\normalsize{%
  \setlength\abovedisplayskip{10pt}
  \setlength\belowdisplayskip{10pt}
}
\newcommand{\commentout}[1]{}

\usepackage{natbib}

\usepackage{algorithm}
\usepackage{algorithmic, tikz, framed}

\usepackage{hyperref,url}

\usepackage{cs886} 

\cstitlerunning{CS886 Project Report}

\begin{document} 

\twocolumn[
\cstitle{Yelp Dataset Challenge: Review Rating Prediction}

% It is OKAY to include author information
\csauthor{Nabiha Asghar}{nasghar@uwaterloo.ca}
\csaddress{University of Waterloo,
            200 University Avenue West, Waterloo, ON N2L3G1 Canada}

\cskeywords{review rating prediction, sentiment analysis, bag of words, bag of opinions, rating prediction, multi-class classification, regression}

\vskip 0.3in
]

\begin{abstract} 
Review websites, such as TripAdvisor and Yelp, allow users to post online reviews for various businesses, products and services, and have been recently shown to have a significant influence on consumer shopping behaviour. An online review typically consists of free-form text and a star rating out of 5.  The problem of predicting a user's star rating for a product, given the user's text review for that product, is called Review Rating Prediction and has lately become a popular, albeit hard, problem in machine learning. In this paper, we treat Review Rating Prediction as a multi-class classification problem, and build sixteen different prediction models by combining four feature extraction methods, (i) unigrams, (ii) bigrams, (iii) trigrams and (iv) Latent Semantic Indexing, with four machine learning algorithms, (i) logistic regression, (ii) Na{\"i}ve Bayes classification, (iii) perceptrons, and (iv) linear Support Vector Classification.  We analyse the performance of each of these sixteen models to come up with the best model for predicting the ratings from reviews. We use the dataset provided by Yelp for training and testing the models. 
\end{abstract} 

\section{Introduction}
\label{problemdesc}

User reviews are an integral part of web services like TripAdvisor, Amazon, Epinions and Yelp, where users can post their opinions about businesses, products and services through reviews consisting of free-form text and a numeric star rating, usually out of 5. These online reviews function as the `online word-of-mouth' \cite{della03} and a criterion for consumers to choose between similar products. Studies (e.g. \cite{chen03}) show that they have a significant impact on consumer purchase decisions as well as on product sales and business revenues. A user review typically looks like this:

\begin{figure}[ht!]
\begin{framed}
\textit{Restaurant: XYZ, Kitchener, N2G4Z6, Canada\\ Rating: \tikz\draw[black,fill=black] (0,0) circle (.5ex); \tikz\draw[black,fill=black] (0,0) circle (.5ex); \tikz\draw[black,fill=black] (0,0) circle (.5ex); \tikz\draw[black] (0,0) circle (.5ex); \tikz\draw[black] (0,0) circle (.5ex);\\ I've been to XYZ a bunch of times. It's a decent place. Nice food, lots of variety! The place is really small though, so you almost never find a spot to sit and eat. The service is also slow at times. }
\end{framed}
\caption{A Typical User Review: Free-form Text \& a Star Rating}
\end{figure}

On famous websites like Amazon and Yelp, many products and businesses receive tens or hundreds of reviews, making it impossible for readers to read all of them. Generally, readers prefer to look at the star ratings only and ignore the text. However, the relationship between the text and the rating is not obvious, as illustrated in Figure 1. In particular, several questions may be asked: why \textit{exactly} did this reviewer give the restaurant 3/5 stars? In addition to the quality of food, variety, size and service time, what other features of the restaurant did the user  implicitly consider, and what was the relative importance given to each of them? How does this relationship change if we consider a different user's rating and text review? 

The process of predicting this relationship for a generic user (but for a specific product/business) is called Review Rating Prediction. Concretely, given the set $S = \{(r_1, s_1), ..., (r_N, s_N)\}$ for a product $P$, where $r_i$ is the $i$'th user's text review of $P$ and $s_i$ is the $i$'th user's numeric rating for $P$, the goal is to learn the best mapping from a word vector \textbf{r} to a numeric rating $s$. Review Rating Prediction is a useful problem to solve, because it can help us decide whether it is enough to look at the star ratings of a product and ignore its textual reviews. Moreover, some review websites allow users to write text reviews without specifying a star rating.  In these cases, Review Rating Prediction comes in handy. However, it is a hard problem because two users who give a product the same rating, may have very different reasons for doing so. User A may give a restaurant 2/5 stars because it does not have free wifi and free parking, even though  the food is good. User B may give the same restaurant a rating of 2/5 because he does not care about the wifi and parking, and thinks that the food is below average. Therefore, the main challenge in building a good predictor is to effectively extract useful features of the product from the text reviews and to then quantify their relative importance with respect to the rating.  

In this paper, we treat Review Rating Prediction Problem as a multi-class classification problem in Machine Learning, where the class labels are the star ratings. We combine four feature extraction methods, unigrams, bigrams, trigrams and Latent Semantic Indexing, with four supervised learning algorithms, logistic regression, Na{\"i}ve Bayes classification, perceptrons and linear support vector classification to build sixteen prediction models. We train and evaluate the performance of each of these models on the dataset provided by Yelp. The rest of the paper is organized as follows. Section 2 and 3 provide the details of the related work and the dataset. Section 4 describes all the feature extraction methods and supervised learning algorithms, and section 6 provides detailed results and analysis. We end with concluding remarks and future work.

\section{Related Work}

Most of the recent work related to review rating prediction relies on sentiment analysis to extract features from the review text. Qu \textit{et al.} \yrcite{qu10} tackle this problem for Amazon.com reviews, by proposing  a novel feature extraction method called bag-of-opinions, which extracts opinions (consisting of a root word, a modifier and/or a negation word) from the review corpus, computes their sentiment score, and predicts a review's rating by aggregating the scores of opinions present in that review and combining it with a domain-dependent unigrams model.  

Leung \textit{et al.} \yrcite{leung06} use a novel \textit{relative frequency method} to create an opinion dictionary, in order to infer review ratings from the review text. This method estimates the strength of a word with respect to a certain sentiment class as the relative frequency of its occurence in that class. They integrate this inference technique with collaborative filtering algorithms and test their method on movie reviews from IMDb on a 2-point rating scale. 

Fan and Khademi \yrcite{fan14} predict a restaurant's average star rating on Yelp from its reviews (note that this is business rating prediction, and is different from review rating prediction). They combine the unigrams model with feature engineering methods such as Parts-of-Speech tagging, and use linear regression, support vector regression and decision trees for prediction. Their dataset consists of 4243 restaurants and 35645 text reviews, and is much smaller than the one we use in this paper. Li \textit{et al.} \yrcite{li11} perform rating prediction for reviews on Epinions.com by extracting additional features of the reviewer and the product/business being reviewed. Ganu \textit{et al.} \yrcite{ganu09} propose a method to use the text of the reviews to improve recommendor systems, like the ones used by Netflix, which often rely solely on the structured metadata information of the product/business and the star ratings. Their method relies on machine learning, sentiment analysis techniques and natural language processing to classify sentences as positive, negative, neutral or conflict. It is shown, using restaurant reviews from Citysearch New York,  that the review text is a better indicator of the sentiment of the review than the coarse star rating. %Sun \textit{et al.}\footnote{http://newport.eecs.uci.edu/$\sim$xis2/Yelp/Final-report-pp16.pdf}

In this paper, we concern ourselves only with the semantic analysis of the review text and do not deal with the sentiment analysis.

\section{Data Description}
\label{datadesc}
We use the dataset provided by Yelp as part of their Dataset Challenge 2014 \cite{yelp14} for training and testing the prediction models. The dataset includes data from Phoenix, Las Vegas, Madison, Waterloo and Edinburgh, and contains information about  42,153 businesses, 320,002 business attributes, 31,617 check-in sets, 403,210 tips and 1,125,458 text reviews. 

Concretely, the dataset consists of five files, one for each object type: business, review, user, check-in and tip. Each file consists of one json-object-per-line. Thus, a business is represented in the `business.json' file as a json object which specifies the business ID, its name, location, stars, review count, opening hours, etc. A text review is a json object in the `review.json' file, which specifies the business ID, user ID, stars (integer values between and including 1 and 5), review text, date and votes.  The necessary data is contained in the business.json and review.json files, therefore we do not use the rest of the data. 

The businesses described in  the Yelp dataset belong to different categories, such as restaurants, shopping, hotels and travel, etc. The text reviews for different business categories may be very different. For example, a typical hotel review may contain the words/phrases `fridge', `television' and  `comfortable bed', but these words would not occur in a restaurant review. Therefore, it is important to perform Review Rating Prediction for each business category independently. That is, the model training and testing for each category should be separate. Figure 2(a) shows the distribution of  business categories in the dataset. Restaurants make up almost 34\% of the 42,153 businesses. Moreover, 68.3\% of the 1,125,458 text reviews are about restaurants, as shown by Figure 2(b). Therefore, in this paper, we restrict ourselves to Review Rating Prediction for restaurants only. Thus, the trimmed dataset that we use consists of 14,403 restaurants and 706,646 reviews. Figure 2(c) shows that the star ratings (out of 5) for the restaurant reviews are not uniformly distributed. About 66\% of these reviews rate the corresponding restaurants very highly (at least 4 stars); the other classes are smaller.

%%%%%%%%%%%%%%%%%%%%%%%%%%%%%%%%%%%%%%%%%%%%%%%%%%%%%%%%%%%%%%%%%%%%%%%%%%%%%%%%%%%%%%%%%%%%%%%%%%%%
%%%%%%%%%%%%%%%%%%%%%%%%%%%%%%%%%%%%%%%%%%%%%%%%%%%%%%%%%%%%%%%%%%%%%%%%%%%%%%%%%%%%%%%%%%%%%%%%%%%%
%%%%%%%%%%%%%%%%%%%%%%%%%%%%%%%%%%%%%%%%%%%%%%%%%%%%%%%%%%%%%%%%%%%%%%%%%%%%%%%%%%%%%%%%%%%%%%%%%%%%
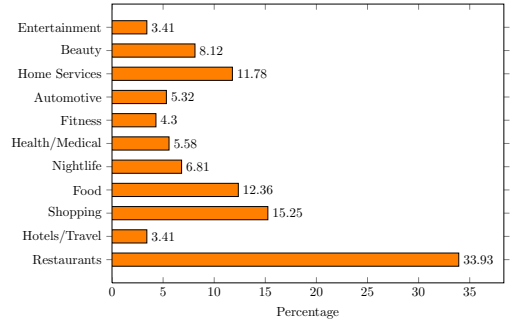
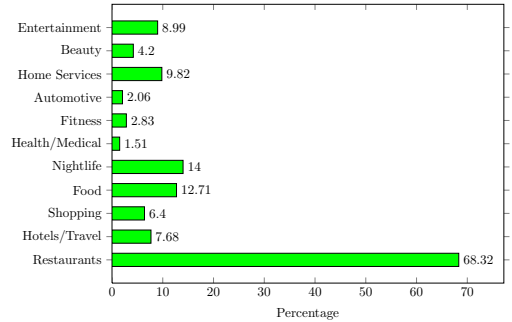
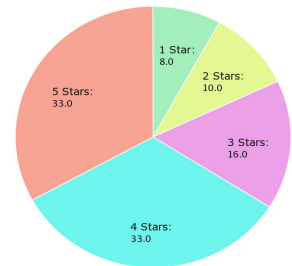
\begin{figure}[ht!]
\centering
\subfigure[Distribution of Business Categories]{
\begin{tikzpicture}[scale=0.5]
\begin{axis}[
    xbar,
    xmin=0.0,
    width=12cm,
    height=9cm,
    enlarge x limits={rel=0.13,upper},
    yticklabels={{Restaurants},{Hotels/Travel},{Shopping},{Food},{Nightlife},{Health/Medical},{Fitness},{Automotive},{Home Services},{Beauty},{Entertainment}},
    xlabel={Percentage},
    ytick=data,
    nodes near coords,
    nodes near coords align=horizontal
]
\addplot [draw=black, fill=orange] coordinates {
    (33.93,1)
    (3.41,2)
    (15.25,3)
   (12.36, 4)
   (6.81, 5)
   (5.58, 6)
   (4.3, 7)
   (5.32, 8)
   (11.78, 9)
   (8.12, 10)
   (3.41, 11)
};
\end{axis}
\end{tikzpicture}
}\\
\subfigure[Distribution of Reviews]{
\begin{tikzpicture}[scale=0.5]
\begin{axis}[
    xbar,
    xmin=0.0,
    width=12cm,
    height=9cm,
    enlarge x limits={rel=0.13,upper},
    yticklabels={{Restaurants},{Hotels/Travel},{Shopping},{Food},{Nightlife},{Health/Medical},{Fitness},{Automotive},{Home Services},{Beauty},{Entertainment}},
    xlabel={Percentage},
    ytick=data,
    nodes near coords,
    nodes near coords align=horizontal
]
\addplot [draw=black, fill=green] coordinates {
    (68.32,1)
    (7.68,2)
    (6.4,3)
   (12.71, 4)
   (14, 5)
   (1.51, 6)
   (2.83, 7)
   (2.06, 8)
   (9.82, 9)
   (4.2, 10)
   (8.99, 11)
};
\end{axis}
\end{tikzpicture}
}\\
\subfigure[Stars Distribution for Restaurant Reviews]{
\includegraphics[width=40mm, height=35mm]{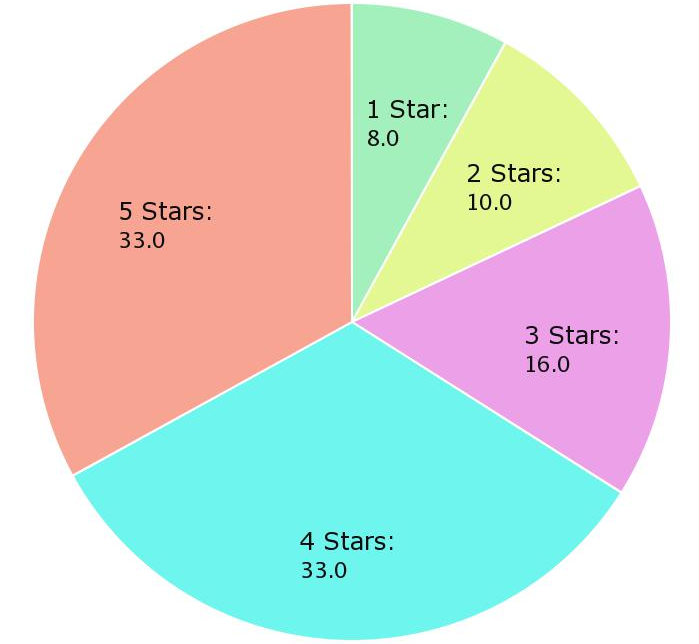}
}
\caption{Descriptive Stats: Yelp Businesses \& Reviews}
\end{figure}

%%%%%%%%%%%%%%%%%%%%%%%%%%%%%%%%%%%%%%%%%%%%%%%%%%%%%%%%%%%%%%%%%%%%%%%%%%%%%%%%%%%%%%%%%%%%%%%%%%%%%
%%%%%%%%%%%%%%%%%%%%%%%%%%%%%%%%%%%%%%%%%%%%%%%%%%%%%%%%%%%%%%%%%%%%%%%%%%%%%%%%%%%%%%%%%%%%%%%%%%%%%
%%%%%%%%%%%%%%%%%%%%%%%%%%%%%%%%%%%%%%%%%%%%%%%%%%%%%%%%%%%%%%%%%%%%%%%%%%%%%%%%%%%%%%%%%%%%%%%%%%%%%

\section{Experimental Setup}
\label{experimentalsetup}

In this paper, we build sixteen different prediction models, by combining each of four different feature extraction methods with each of four distinct supervised learning algorithms. In this section, we describe the preprocessing phase, the four feature extraction methods, the four supervised learning algorithms, and two performance evaluation metrics. 
\subsection{Preprocessing}
We first write some basic Python scripts to separate the restaurants from the business.json file, and to separate the restaurant reviews from the review.json file. We then preprocess the text reviews as follows.

Yelp allows users to write text reviews in free form. This means that a user may excessively use capital letters and punctuation marks (to express his/her intense dislike, for example) and slang words within a review. Moreover, stop words, like `the', `that', `is' etc, occur frequently across reviews and are not very useful. Therefore it is necessary to preprocess the reviews in order to extract meaningful content from each of them.  To do this, we use standard Python libraries to remove capitalizations, stop words and punctuations. %We also tokenize the text, using Stanford Tokenizer\footnote{Stanford's open source NLP software is available at http://nlp.stanford.edu/software/index.shtml.}, in order to identify the nouns, adjectives and other constructs within in each sentence.

\subsection{Feature Extraction}

We use four methods to extract useful features from the review corpus and to build a feature vector for each review. Each of these methods relies on semantic analysis of the text.
\subsubsection{Unigrams}
In the uni-grams model (also called the ``bag of words" model), each unique word in the pre-processed review corpus is considered as a feature. Thus, building a feature vector for a review is straightforward. First, a dictionary of all the words occuring in the review corpus is created. Then a word-review matrix is constructed, where entry $(i,j)$ is the frequency of occurence of word $i$ in the $j$'th review.  Finally, we apply the TF-IDF (Term Frequency -  Inverse Document Frequency) weighting technique to this matrix to obtain the final feature matrix. This weighting technique assigns less weight to words that occur more frequently across reviews (e.g. ``food") because they are generally not good distinguishers between any pair of reviews, and a high weight to more rare words.  Each column of this matrix is a feature vector of the corresponding review.

The size of the dictionary in this case, i.e. the total number of features, is 171,846.
\subsubsection{Unigrams \& Bigrams}

The unigrams model is a widely used feature extraction method in natural language processing. It is quite simple to implement and in many cases, it gives surprisingly good results.  However, its inherent drawback is its inability to capture relationships between two words (e.g. a word and its modifier, a word and its negation, etc), because it treats each word in isolation. To capture the effect of phrases such as `tasty burger' and `not delicious', we \textit{add} bigrams to the unigrams model. Now, the dictionary additionally consists of all the 2-tuples of words (i.e. all pairs of consecutive words) occuring in the corpus of reviews.  The matrix is computed as before; it has more rows now. As before, we apply TF-IDF weighting to this matrix so that less importance is given to common words and more importance is given to rare words.

The size of the dictionary (total number of features) is 7,612,422.  
\subsubsection{Unigrams, Bigrams \& Trigrams}
To capture the effect of phrases like `tasty fish burger', we now add trigrams (i.e. all triples of consecutive words) to the unigrams+bigrams model. The rest of the computations for building the feature matrix are the same as before. Note, however, that the same trigram would rarely occur across different reviews, because two different people are unlikely to use the same 3-word phrase in their reviews. Therefore,  the results of this model are not expected to be very different from the unigrams+bigrams model. 

The total number of features in this case is 31,677,669.
% Moreover, a Part-of-Speech Tagger\footnote{http://nlp.stanford.edu/software/index.shtml}  may be used to extract meaningful adjective-noun phrases (such as \textit{helpful website, nice food}), which can be used as additional features.
\subsubsection{Latent Semantic Indexing (LSI)}
Latent Semantic Indexing (LSI) \cite{hofmann99} is a more sophisticated method of lexical matching, which goes beyond exact matching of words. It finds `topics' in reviews, which are words having similar meanings or words occuring in a similar context. In LSI, we first construct a word-review matrix $M$, of size $m \times t$, using the unigrams model, and then do Singular Value Decomposition (SVD) of $M$.
\begin{equation*}
SVD(M) = U \cdot S \cdot V^T
\end{equation*}
The SVD function outputs three matrices: the word-topic matrix $U$ of size $m \times m$, the rectangular diagonal matrix $S$ of size $m \times t$ containing $t$ singular values, and the transpose of the topic-review matrix $V$ of size $t \times t$. We use $V$ as the feature matrix.

The singular values matrix $S$ has $t$ non-zero diagonal entries that are the singular values in decreasing order of importance. The columns of $S$ correspond to the topics in the reviews. The $i$'th singular value is a measure of the importance of the $i$'th topic.  By default, $t$ equals the size of vocabulary (i.e. 171,846). However, the first $t^*$ topics can be chosen as the most important ones, and thus the top $t^*$ rows of $V$ can be used as the feature matrix. Determining the value of $t^*$ is crucial, and this can be done by examining a simple plot of the singular values against their importance, and looking for an `elbow' in the plot.

\subsection{Supervised Learning} 
To train our prediction models, we use four supervised learning algorithms. 
\subsubsection{Logistic Regression}
In logistic regression \cite{freedman09}, the conditional probability function $P(s|$\textbf{r}) is modelled, where \textbf{r} is a feature vector for review $r$ and $s$ belongs to the set of class labels $ \{1,2,3,4,5\}$. Then, given a new feature vector $\textbf{r}^*$ for a new review $r^*$, this probability function is computed for all values of $s$, and the $s$ value corresponding to the highest probability is output as the final class label (star rating) for this review.

\subsubsection{Na{\"i}ve Bayes Classification}
A Na{\"i}ve Bayes \cite{ng02} classifier makes the Na{\"i}ve Bayes assumption (i.e. it assumes conditional independence between any pair of features given some class) to model the joint probability $P($\textbf{r}$,s)$ for any feature vector \textbf{r} and star rating $s$. Then, given a new feature vector $\textbf{r}^*$ for a new review $r^*$, the joint probability function is computed for all values of $s$, and the $s$ value corresponding to the highest probability is output as the final class label for review $r^*$.

In this paper, we use multinomial Na{\"i}ve Bayes classification, which assumes that  $P($\textbf{r}$_i |s)$ is a multinomial distribution for all $i$. This is a typical choice for document classification, because it works well for data that can be turned into counts, for example  weighted word frequencies in the text.

\subsubsection{Perceptrons}
A perceptron \cite{rosenblatt57} is a linear classifier that outputs class labels instead of probabilities. It uses a gradient-descent-like rule to iterate over the training set multiple times, in order to re-classify any misclassified examples, until all of them have been classified correctly. For linearly separable data, a Perceptron Convergence Rule states that a solution will always be found after some finite number of iterations. For data that is not linearly separable, there will be oscillation, which can be detected automatically. 

Perceptron solutions may be non-unique, because the margin of the linear decision boundary  is ignored. In our experiments, we set the number of iterations to be 50 (a typical choice), that is, the classifier loops over the entire training set 50 times. 

\subsubsection{Linear Support Vector Classification (SVC)}
Support Vector Machines (SVM) \cite{tsochantaridis04} are enhanced versions of perceptrons, in that they eliminate the non-uniqueness of solutions by optimizing the margin around the decision boundary, and handle non-separable data by allowing misclassifications. A parameter $C$ controls overfitting. When $C$ is small, the algorithm focuses on maximizing the margin, even if this means more misclassifications, and for lage values of $C$, the margin is decreased if this helps to classify more examples correctly. 

In our experiments, we use linear SVMs for multi-class classification. The tolerance of the convergence criterion is set to 0.001. For each feature extraction method, we do internal 3-fold cross validation to choose the value of $C$ that gives the highest accuracy. It turns out that $C=1.0$ works best every time.

\subsection{Performance Metrics \& Implementation Details}

We use 80\% of the dataset for training, and 20\% for testing. For each of the sixteen prediction systems, we perform 3-fold cross validation on the training set and compute two metrics, Root Mean Squared Error (RMSE) and accuracy, for the training fold as well as the validation fold.

All the implementation is done on an Intel Core i5 CPU with 4 cores (1.6 GHz each), 8 GB RAM and 64 bit Ubuntu 12.04 operating system. The programming language used is Python, and extensive use is made of its libraries numpy, scipy and scikit-learn. 

\section{Results and Analysis}

In this section, we present the results for the four feature extraction methods separately. For each method, we show an RMSE graph and an Accuracy graph; each graph contains plots for the four classifiers. We then analyze these results to choose the best of the sixteen systems, and finally we evaluate the chosen system on the test set.

\subsection{Unigrams}
Figures 3(a) and 4(a) show the performance of the four classifiers on unigrams' features. The total number of available features is 171,846 and we do not know how many of these are useful, so we plot the RMSE and the accuracy against the top\footnote{For example, top 10 features would be the features with the top 10 TF-IDF weights.} $x$ number of features. We see that as the number of features increases, the training-fold RMSE for each classifier increases and the training-fold accuracy of each classifier decreases, but the validation-fold RMSE and the validation-fold accuracy level off at about 10,000 features. The only exception is the Na{\"i}ve Bayes classifier, whose RMSE reaches a minimum value around 10,000 features, but then starts rising again. 

Clearly, perceptrons perform the worst, achieving a lowest RMSE of 1.25 and the highest accuracy of 43\%. The Na{\"i}ve Bayes classifier is the second worst, with the best RMSE and accuracy values of  0.96 and 52\%. The performances of Linear SVC and logistic regression are not significantly different on the validation fold. Linear SVC's best RMSE and accuracy scores are 0.87 and 57\% , while those for logistic regression are 0.85 and 58\%. So, logistic regression has the best performance for unigrams. 

Note that a random (coin-flip) classifier, that assumes no knowledge of the data distribution, would have an accuracy of 20\%, because we have 5 classes. Logistic regression improves the random classifier accuracy by almost 300\%.
%%%%%%%%%%%%%%%%%%%%%%%%%%%%%%%%%%%%%%%%%%%%%%%%%%%%%%%%%%%%%%%%%%%%%%%%%%%%%%%%%%%%%%%%%%%%%%%%%%%%%%%%%%%]
%%%%%%%%%%%%%%%%%%%%%%%%%%%%%%%%%%%%%%%%%%%%%%%%%%%%%%%%%%%%%%%%%%%%%%%%%%%%%%%%%%%%%%%%%%%%%%%%%%%%%%%%%%%]
%%%%%%%%%%%%%%%%%%%%%%%%%%%%%%%%%%%%%%%%%%%%%%%%%%%%%%%%%%%%%%%%%%%%%%%%%%%%%%%%%%%%%%%%%%%%%%%%%%%%%%%%%%%]
%%%%%%%%%%%%%%%%%%%%%%%%%%%%%%%%%%%%%%%%%%%%%%%%%%%%%%%%%%%%%%%%%%%%%%%%%%%%%%%%%%%%%%%%%%%%%%%%%%%%%%%%%%%]

\begin{figure*}[ht!]
\centering
\subfigure[Unigrams]{
\begin{tikzpicture}[scale=0.8]
\begin{axis}[
legend style={legend pos=north east, font=\tiny},
xlabel={No. of Features},
ylabel={RMSE},
xmin=-500, xmax=60000,
ymin=0.6, ymax=1.7,
xtick={0, 10000, 20000, 30000, 40000, 50000, 60000, 70000},
ytick={0.6,0.7, 0.8, 0.9, 1.0, 1.1, 1.2, 1.3, 1.4, 1.5, 1.6, 1.7},
ymajorgrids=true,
grid style=dashed
]
\addplot[
%dash pattern=on 4pt off 1pt on 4pt off 4pt,
%dashed,
color=red,
%mark=*,
very thick,
error bars,
y dir= both,
y explicit
]
coordinates {
(20, 1.47)
(40, 1.375)
(60, 1.29)
(80, 1.27)
(100, 1.2)
(200, 1.13)
(300, 1.077)
(400, 1.04)
(500, 1.025)
(600, 0.98)
(700, 0.96)
(800, 0.95)
(900, 0.94)
(1000, 0.927)
(2000, 0.88)
(3000, 0.866)
(4000, 0.86)
(5000, 0.854)
(6000, 0.854)
(7000, 0.854)
(8000, 0.85)
(9000, 0.85)
(10000, 0.85)
(15000, 0.85)
(20000, 0.85)
(30000, 0.85)
(40000, 0.85)
(50000, 0.85)
(55000, 0.85)
(60000, 0.85)
(65000, 0.85)
(70000, 0.85)
(75000, 0.85)
(80000, 0.85)
(90000, 0.85)
(100000, 0.85)
};\addlegendentry{Log Reg val}
\addplot[
%dash pattern=on 4pt off 1pt on 4pt off 4pt,
dashed,
color=red,
%mark=*,
very thick,
error bars,
y dir= both,
y explicit
]
coordinates {
(20, 1.47)
(40, 1.37)
(60, 1.3)
(80, 1.26)
(100, 1.21)
(200, 1.12)
(300, 1.07)
(400, 1.03)
(500, 1.01)
(600, 0.97)
(700, 0.95)
(800, 0.94)
(900, 0.927)
(1000, 0.85)
(2000, 0.866)
(3000, 0.85)
(4000, 0.8366)
(5000, 0.83)
(6000, 0.82)
(7000, 0.82)
(8000, 0.82)
(9000, 0.81)
(10000, 0.81)
(15000, 0.8)
(20000, 0.8)
(30000, 0.79)
(40000, 0.79)
(50000, 0.787)
(60000, 0.787)
};\addlegendentry{Log Reg Train}
\addplot[
%dash pattern=on 4pt off 1pt on 4pt off 4pt,
%dashed,
color=yellow,
%mark=*,
very thick,
error bars,
y dir= both,
y explicit
]
coordinates {
(20, 1.4)
(40, 1.36)
(60, 1.33)
(80, 1.32)
(100, 1.28)
(200, 1.24)
(300, 1.2)
(400, 1.18)
(500, 1.18)
(600, 1.12)
(700, 1.1)
(800, 1.086)
(900, 1.077)
(1000, 1.067)
(2000, 1.01)
(3000, 0.99)
(4000, 0.98)
(5000, 0.97)
(6000, 0.97)
(7000, 0.97)
(8000, 0.96)
(9000, 0.96)
(10000, 0.96)
(15000, 0.96)
(20000, 0.97)
(30000, 0.98)
(40000, 1.0)
(50000, 1.02)
(55000, 1.02)
(60000, 1.034)
(65000, 1.044)
(70000, 1.05)
(75000, 1.063)
(80000, 1.07)
(90000, 1.09)
(100000, 1.11)
};\addlegendentry{NB val}
\addplot[
%dash pattern=on 4pt off 1pt on 4pt off 4pt,
dashed,
color=yellow,
%mark=*,
very thick,
error bars,
y dir= both,
y explicit
]
coordinates {
(20, 1.4)
(40, 1.36)
(60, 1.338)
(80, 1.32)
(100, 1.28)
(200, 1.237)
(300, 1.2)
(400, 1.175)
(500, 1.17)
(600, 1.11)
(700, 1.095)
(800, 1.086)
(900, 1.077)
(1000, 1.063)
(2000, 1.0)
(3000, 0.98)
(4000, 0.97)
(5000, 0.96)
(6000, 0.96)
(7000, 0.95)
(8000, 0.95)
(9000, 0.95)
(10000, 0.95)
(15000, 0.94)
(20000, 0.94)
(30000, 0.95)
(40000, 0.97)
(50000, 0.98)
(60000, 1.0)
};\addlegendentry{NB train}
\addplot[
%dash pattern=on 4pt off 1pt on 4pt off 4pt,
%dashed,
color=green,
%mark=*,
very thick,
error bars,
y dir= both,
y explicit
]
coordinates {
(20, 1.706)
(40, 1.67)
(60, 1.66)
(80, 1.62)
(100, 1.63)
(200, 1.536)
(300, 1.49)
(400, 1.47)
(500, 1.47)
(600, 1.43)
(700, 1.42)
(800, 1.41)
(900, 1.4)
(1000, 1.36)
(2000, 1.32)
(3000, 1.31)
(4000, 1.3)
(5000, 1.28)
(6000, 1.257)
(7000, 1.26)
(8000, 1.26)
(9000, 1.26)
(10000, 1.26)
(15000, 1.26)
(20000, 1.25)
(30000, 1.25)
(40000, 1.26)
(50000, 1.25)
(55000, 1.257)
(60000, 1.25)
(65000, 1.257)
(70000, 1.253)
(75000, 1.253)
(80000, 1.26)
(90000, 1.26)
(100000, 1.257)
};\addlegendentry{Perc val}
\addplot[
%dash pattern=on 4pt off 1pt on 4pt off 4pt,
dashed,
color=green,
%mark=*,
very thick,
error bars,
y dir= both,
y explicit
]
coordinates {
(20, 1.65)
(40, 1.603)
(60, 1.55)
(80, 1.578)
(100, 1.45)
(200, 1.4177)
(300, 1.304)
(400, 1.288)
(500, 1.27)
(600, 1.22)
(700, 1.208)
(800, 1.19)
(900, 1.166)
(1000, 1.175)
(2000, 1.11)
(3000, 1.086)
(4000, 1.07)
(5000, 1.07)
(6000, 1.05)
(7000, 1.044)
(8000, 1.034)
(9000, 1.03)
(10000, 1.02)
(15000, 1.0)
(20000, 0.98)
(30000, 0.96)
(40000, 0.95)
(50000, 0.94)
(60000, 0.927)
};\addlegendentry{Perc train}
\addplot[
%dash pattern=on 4pt off 1pt on 4pt off 4pt,
%dashed,
color=blue,
%mark=*,
very thick,
error bars,
y dir= both,
y explicit
]
coordinates {
(20, 1.476)
(40, 1.39)
(60, 1.315)
(80, 1.29)
(100, 1.24)
(200, 1.16)
(300, 1.1)
(400, 1.063)
(500, 1.05)
(600, 1.0)
(700, 0.98)
(800, 0.964)
(900, 0.95)
(1000, 0.94)
(2000, 0.89)
(3000, 0.88)
(4000, 0.87)
(5000, 0.87)
(6000, 0.866)
(7000, 0.866)
(8000, 0.87)
(9000, 0.87)
(10000, 0.87)
(15000, 0.87)
(20000, 0.877)
(30000, 0.88)
(40000, 0.88)
(50000, 0.88)
(55000, 0.88)
(60000, 0.88)
(65000, 0.88)
(70000, 0.88)
(75000, 0.88)
(80000, 0.88)
(90000, 0.88)
(100000, 0.88)
};\addlegendentry{Lin SVC val}
\addplot[
%dash pattern=on 4pt off 1pt on 4pt off 4pt,
dashed,
color=blue,
%mark=*,
very thick,
error bars,
y dir= both,
y explicit
]
coordinates {
(20, 1.476)
(40, 1.39)
(60, 1.323)
(80, 1.29)
(100, 1.24)
(200, 1.16)
(300, 1.1)
(400, 1.095)
(500, 1.04)
(600, 0.99)
(700, 0.97)
(800, 0.96)
(900, 0.95)
(1000, 0.93)
(2000, 0.88)
(3000, 0.86)
(4000, 0.85)
(5000, 0.837)
(6000, 0.83)
(7000, 0.82)
(8000, 0.82)
(9000, 0.81)
(10000, 0.806)
(15000, 0.79)
(20000, 0.77)
(30000, 0.75)
(40000, 0.74)
(50000, 0.73)
(60000, 0.72)
};\addlegendentry{Lin SVC train}
\end{axis}
\end{tikzpicture}
}
\subfigure[Unigrams \& Bigrams]{
\begin{tikzpicture}[scale=0.8]
\begin{axis}[
legend style={legend pos=north east, font=\tiny},
xlabel={No. of Features},
ylabel={RMSE},
xmin=-500, xmax=60000,
ymin=0.6, ymax=1.7,
xtick={0, 10000, 20000, 30000, 40000, 50000, 60000, 70000},
ytick={0.6,0.7, 0.8, 0.9, 1.0, 1.1, 1.2, 1.3, 1.4, 1.5, 1.6,1.7},
ymajorgrids=true,
grid style=dashed
]
\addplot[
%dash pattern=on 4pt off 1pt on 4pt off 4pt,
%dashed,
color=red,
%mark=*,
very thick,
error bars,
y dir= both,
y explicit
]
coordinates {
(20, 1.47)
(40, 1.37)
(60, 1.28)
(80, 1.27)
(100, 1.19)
(200, 1.1)
(300, 1.07)
(400, 1.04)
(500, 1.025)
(600, 0.94)
(700, 0.91)
(800, 0.89)
(900, 0.85)
(1000, 0.84)
(2000, 0.82)
(3000, 0.80)
(4000, 0.79)
(5000, 0.79)
(6000, 0.79)
(7000, 0.78)
(8000, 0.78)
(9000, 0.78)
(10000, 0.78)
(15000, 0.78)
(20000, 0.78)
(30000, 0.78)
(40000, 0.78)
(50000, 0.78)
(55000, 0.78)
(60000, 0.78)
};\addlegendentry{Log Reg val}
\addplot[
%dash pattern=on 4pt off 1pt on 4pt off 4pt,
dashed,
color=red,
%mark=*,
very thick,
error bars,
y dir= both,
y explicit
]
coordinates {
(20, 1.47)
(40, 1.37)
(60, 1.3)
(80, 1.24)
(100, 1.17)
(200, 1.09)
(300, 1.05)
(400, 1.02)
(500, 0.99)
(600, 0.95)
(700, 0.92)
(800, 0.91)
(900, 0.87)
(1000, 0.84)
(2000, 0.80)
(3000, 0.78)
(4000, 0.77)
(5000, 0.76)
(6000, 0.75)
(7000, 0.75)
(8000, 0.74)
(9000, 0.73)
(10000, 0.72)
(15000, 0.71)
(20000, 0.70)
(30000, 0.68)
(40000, 0.68)
(50000, 0.67)
(60000, 0.67)
};\addlegendentry{Log Reg Train}
\addplot[
%dash pattern=on 4pt off 1pt on 4pt off 4pt,
%dashed,
color=yellow,
%mark=*,
very thick,
error bars,
y dir= both,
y explicit
]
coordinates {
(20, 1.4)
(40, 1.36)
(60, 1.33)
(80, 1.32)
(100, 1.28)
(200, 1.24)
(300, 1.2)
(400, 1.18)
(500, 1.18)
(600, 1.12)
(700, 1.05)
(800, 1.01)
(900, 0.99)
(1000, 0.95)
(2000, 0.93)
(3000, 0.92)
(4000, 0.91)
(5000, 0.90)
(6000, 0.89)
(7000, 0.88)
(8000, 0.88)
(9000, 0.88)
(10000, 0.89)
(15000, 0.89)
(20000, 0.90)
(30000, 0.91)
(40000, 0.92)
(50000, 0.92)
(55000, 0.94)
(60000, 0.95)
};\addlegendentry{NB val}
\addplot[
%dash pattern=on 4pt off 1pt on 4pt off 4pt,
dashed,
color=yellow,
%mark=*,
very thick,
error bars,
y dir= both,
y explicit
]
coordinates {
(20, 1.4)
(40, 1.36)
(60, 1.338)
(80, 1.32)
(100, 1.28)
(200, 1.237)
(300, 1.1)
(400, 1.04)
(500, 0.98)
(600, 0.97)
(700, 0.96)
(800, 0.95)
(900, 0.94)
(1000, 0.93)
(2000, 0.92)
(3000, 0.91)
(4000, 0.90)
(5000, 0.89)
(6000, 0.87)
(7000, 0.86)
(8000, 0.85)
(9000, 0.84)
(10000, 0.83)
(15000, 0.82)
(20000, 0.82)
(30000, 0.82)
(40000, 0.82)
(50000, 0.83)
(60000, 0.83)
};\addlegendentry{NB train}
\addplot[
%dash pattern=on 4pt off 1pt on 4pt off 4pt,
%dashed,
color=green,
%mark=*,
very thick,
error bars,
y dir= both,
y explicit
]
coordinates {
(20, 1.62)
(40, 1.61)
(60, 1.60)
(80, 1.59)
(100, 1.56)
(200, 1.536)
(300, 1.49)
(400, 1.47)
(500, 1.47)
(600, 1.43)
(700, 1.42)
(800, 1.41)
(900, 1.4)
(1000, 1.36)
(2000, 1.32)
(3000, 1.27)
(4000, 1.23)
(5000, 1.20)
(6000, 1.19)
(7000, 1.19)
(8000, 1.19)
(9000, 1.19)
(10000, 1.18)
(15000, 1.18)
(20000, 1.18)
(30000, 1.18)
(40000, 1.18)
(50000, 1.18)
(55000, 1.18)
(60000, 1.18)
};\addlegendentry{Perc val}
\addplot[
%dash pattern=on 4pt off 1pt on 4pt off 4pt,
dashed,
color=green,
%mark=*,
very thick,
error bars,
y dir= both,
y explicit
]
coordinates {
(20, 1.62)
(40, 1.603)
(60, 1.55)
(80, 1.50)
(100, 1.45)
(200, 1.40)
(300, 1.37)
(400, 1.36)
(500, 1.35)
(600, 1.34)
(700, 1.32)
(800, 1.31)
(900, 1.30)
(1000, 1.28)
(2000, 1.24)
(3000, 1.21)
(4000, 1.18)
(5000, 1.15)
(6000, 1.12)
(7000, 1.09)
(8000, 1.04)
(9000, 1.02)
(10000, 0.97)
(15000, 0.95)
(20000, 0.92)
(30000, 0.89)
(40000, 0.87)
(50000, 0.86)
(60000, 0.83)
};\addlegendentry{Perc train}
\addplot[
%dash pattern=on 4pt off 1pt on 4pt off 4pt,
%dashed,
color=blue,
%mark=*,
very thick,
error bars,
y dir= both,
y explicit
]
coordinates {
(20, 1.42)
(40, 1.36)
(60, 1.315)
(80, 1.29)
(100, 1.24)
(200, 1.16)
(300, 1.1)
(400, 1.063)
(500, 1.05)
(600, 1.0)
(700, 0.98)
(800, 0.964)
(900, 0.91)
(1000, 0.87)
(2000, 0.84)
(3000, 0.82)
(4000, 0.82)
(5000, 0.82)
(6000, 0.82)
(7000, 0.82)
(8000, 0.82)
(9000, 0.81)
(10000, 0.81)
(15000, 0.81)
(20000, 0.81)
(30000, 0.81)
(40000, 0.81)
(50000, 0.81)
(55000, 0.81)
(60000, 0.81)
};\addlegendentry{Lin SVC val}
\addplot[
%dash pattern=on 4pt off 1pt on 4pt off 4pt,
dashed,
color=blue,
%mark=*,
very thick,
error bars,
y dir= both,
y explicit
]
coordinates {
(20, 1.42)
(40, 1.37)
(60, 1.323)
(80, 1.29)
(100, 1.24)
(200, 1.16)
(300, 1.1)
(400, 1.095)
(500, 1.04)
(600, 0.98)
(700, 0.92)
(800, 0.87)
(900, 0.81)
(1000, 0.79)
(2000, 0.77)
(3000, 0.75)
(4000, 0.74)
(5000, 0.737)
(6000, 0.73)
(7000, 0.72)
(8000, 0.71)
(9000, 0.70)
(10000, 0.69)
(15000, 0.67)
(20000, 0.65)
(30000, 0.64)
(40000, 0.63)
(50000, 0.62)
(60000, 0.61)
};\addlegendentry{Lin SVC train}
\end{axis}
\end{tikzpicture}
}
\caption{RMSE plots for (a) Unigrams, and (b) Unigrams \& Bigrams}
\label{}
\end{figure*}
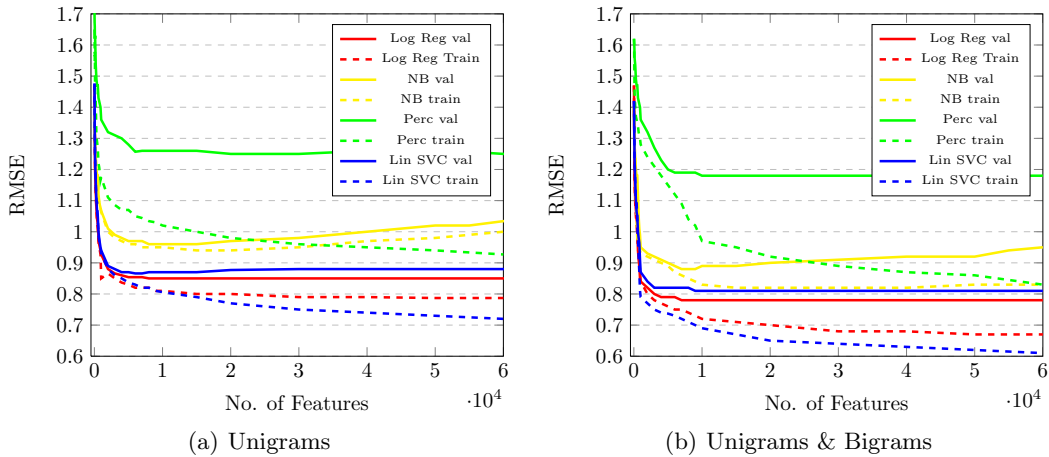
%%%%%%%%%%%%%%%%%%%%%%%%%%%%%%%%%%%%%%%%%%%%%%%%%%%%%%%%%%%%%%%%%%%%%%%%%%%%%%%%
%%%%%%%%%%%%%%%%%%%%%%%%%%%%%%%%%%%%%%%%%%%%%%%%%%%%%%%%%%%%%%%%%%%%%%%%%%%%%%%%
%%%%%%%%%%%%%%%%%%%%%%%%%%%%%%%%%%%%%%%%%%%%%%%%%%%%%%%%%%%%%%%%%%%%%%%%%%%%%%%%
\subsection{Unigrams \& Bigrams}

Figures 3(b) and 4(b) show the performance of the four classifiers on the top $x$ number of features obtained from unigrams \& bigrams. The total number of available features in this case is over 7 million. We see that the RMSE and accuracy improves for every classifier, compared to Figures 3(a) and 4(a). This is  because bigrams occur frequently enough in the corpus, capture a lot of information that unigrams cannot, and give much more meaningful features. However, the overall trends are quite the same for the training and validation folds. Perceptrons are still the worst, followed by the Na{\"i}ve Bayes classifier. Linear SVC and logistic regression are quite close again. Linear SVC's best RMSE and accuracy scores are 0.81 and 63\%, while those for logistic regression are 0.78 and 64\%. Logistic regression wins again. 

%%%%%%%%%%%%%%%%%%%%%%%%%%%%%%%%%%%%%%%%%%%%%%%%%%%%%%%%%%%%%%%%%%%%%%%%%%%%%%%%%%%%%%%%%%%%%%%%%%%%%%%%%%%]
%%%%%%%%%%%%%%%%%%%%%%%%%%%%%%%%%%%%%%%%%%%%%%%%%%%%%%%%%%%%%%%%%%%%%%%%%%%%%%%%%%%%%%%%%%%%%%%%%%%%%%%%%%%]
%%%%%%%%%%%%%%%%%%%%%%%%%%%%%%%%%%%%%%%%%%%%%%%%%%%%%%%%%%%%%%%%%%%%%%%%%%%%%%%%%%%%%%%%%%%%%%%%%%%%%%%%%%%]
%%%%%%%%%%%%%%%%%%%%%%%%%%%%%%%%%%%%%%%%%%%%%%%%%%%%%%%%%%%%%%%%%%%%%%%%%%%%%%%%%%%%%%%%%%%%%%%%%%%%%%%%%%%]

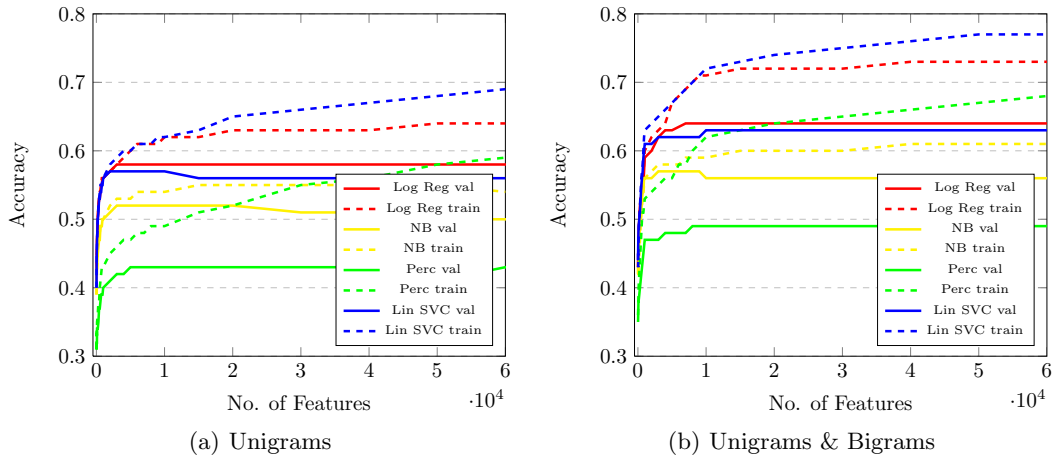
\begin{figure*}[ht!]
\centering
\subfigure[Unigrams]{
\begin{tikzpicture}[scale=0.8]
\begin{axis}[
legend style={legend pos=south east, font=\tiny},
xlabel={No. of Features},
ylabel={Accuracy},
xmin=-500, xmax=60000,
ymin=0.3, ymax=0.8,
xtick={0, 10000, 20000, 30000, 40000, 50000, 60000, 70000},
ytick={0.0, 0.1, 0.2, 0.3, 0.4, 0.5, 0.6, 0.7, 0.8, 0.9, 1.0},
ymajorgrids=true,
grid style=dashed
]
\addplot[
%dash pattern=on 4pt off 1pt on 4pt off 4pt,
%dashed,
color=red,
%mark=*,
very thick,
error bars,
y dir= both,
y explicit
]
coordinates {
(20, 0.41)
(40, 0.43)
(60, 0.45)
(80, 0.46)
(100, 0.47)
(200, 0.50)
(300, 0.52)
(400, 0.53)
(500, 0.53)
(600, 0.54)
(700, 0.55)
(800, 0.55)
(900, 0.56)
(1000, 0.56)
(2000, 0.57)
(3000, 0.58)
(4000, 0.58)
(5000, 0.58)
(6000, 0.58)
(7000, 0.58)
(8000, 0.58)
(9000, 0.58)
(10000, 0.58)
(15000, 0.58)
(20000, 0.58)
(30000, 0.58)
(40000, 0.58)
(50000, 0.58)
(55000, 0.58)
(60000, 0.58)
(65000, 0.58)
(70000, 0.58)
(75000, 0.58)
(80000, 0.58)
(90000, 0.58)
(100000, 0.58)
};\addlegendentry{Log Reg val}
\addplot[
%dash pattern=on 4pt off 1pt on 4pt off 4pt,
dashed,
color=red,
%mark=*,
very thick,
error bars,
y dir= both,
y explicit
]
coordinates {
(20, 0.41)
(40, 0.43)
(60, 0.45)
(80, 0.46)
(100, 0.47)
(200, 0.50)
(300, 0.52)
(400, 0.53)
(500, 0.54)
(600, 0.54)
(700, 0.55)
(800, 0.56)
(900, 0.56)
(1000, 0.56)
(2000, 0.57)
(3000, 0.58)
(4000, 0.59)
(5000, 0.60)
(6000, 0.61)
(7000, 0.61)
(8000, 0.61)
(9000, 0.61)
(10000, 0.62)
(15000, 0.62)
(20000, 0.63)
(30000, 0.63)
(40000, 0.63)
(50000, 0.64)
(60000, 0.64)
};\addlegendentry{Log Reg train}
\addplot[
%dash pattern=on 4pt off 1pt on 4pt off 4pt,
%dashed,
color=yellow,
%mark=*,
very thick,
error bars,
y dir= both,
y explicit
]
coordinates {
(20, 0.39)
(40, 0.40)
(60, 0.42)
(80, 0.42)
(100, 0.43)
(200, 0.45)
(300, 0.46)
(400, 0.47)
(500, 0.47)
(600, 0.49)
(700, 0.49)
(800, 0.49)
(900, 0.5)
(1000, 0.5)
(2000, 0.51)
(3000, 0.52)
(4000, 0.52)
(5000, 0.52)
(6000, 0.52)
(7000, 0.52)
(8000, 0.52)
(9000, 0.52)
(10000, 0.52)
(15000, 0.52)
(20000, 0.52)
(30000, 0.51)
(40000, 0.51)
(50000, 0.5)
(55000, 0.5)
(60000, 0.5)
(65000, 0.49)
(70000, 0.49)
(75000, 0.49)
(80000, 0.49)
(90000, 0.48)
(100000, 0.48)
};\addlegendentry{NB val}
\addplot[
%dash pattern=on 4pt off 1pt on 4pt off 4pt,
dashed,
color=yellow,
%mark=*,
very thick,
error bars,
y dir= both,
y explicit
]
coordinates {
(20, 0.39)
(40, 0.40)
(60, 0.42)
(80, 0.42)
(100, 0.43)
(200, 0.45)
(300, 0.47)
(400, 0.47)
(500, 0.48)
(600, 0.49)
(700, 0.49)
(800, 0.50)
(900, 0.5)
(1000, 0.5)
(2000, 0.52)
(3000, 0.53)
(4000, 0.53)
(5000, 0.53)
(6000, 0.54)
(7000, 0.54)
(8000, 0.54)
(9000, 0.54)
(10000, 0.54)
(15000, 0.55)
(20000, 0.55)
(30000, 0.55)
(40000, 0.55)
(50000, 0.55)
(60000, 0.54)
};\addlegendentry{NB train}
\addplot[
%dash pattern=on 4pt off 1pt on 4pt off 4pt,
%dashed,
color=green,
%mark=*,
very thick,
error bars,
y dir= both,
y explicit
]
coordinates {
(20, 0.32)
(40, 0.31)
(60, 0.32)
(80, 0.32)
(100, 0.34)
(200, 0.34)
(300, 0.35)
(400, 0.37)
(500, 0.37)
(600, 0.38)
(700, 0.39)
(800, 0.39)
(900, 0.39)
(1000, 0.4)
(2000, 0.41)
(3000, 0.42)
(4000, 0.42)
(5000, 0.43)
(6000, 0.43)
(7000, 0.43)
(8000, 0.43)
(9000, 0.43)
(10000, 0.43)
(15000, 0.43)
(20000, 0.43)
(30000, 0.43)
(40000, 0.43)
(50000, 0.43)
(55000, 0.42)
(60000, 0.43)
(65000, 0.42)
(70000, 0.42)
(75000, 0.42)
(80000, 0.42)
(90000, 0.42)
(100000, 0.42)
};\addlegendentry{Perc val}
\addplot[
%dash pattern=on 4pt off 1pt on 4pt off 4pt,
dashed,
color=green,
%mark=*,
very thick,
error bars,
y dir= both,
y explicit
]
coordinates {
(20, 0.31)
(40, 0.32)
(60, 0.34)
(80, 0.33)
(100, 0.35)
(200, 0.36)
(300, 0.38)
(400, 0.38)
(500, 0.40)
(600, 0.41)
(700, 0.42)
(800, 0.43)
(900, 0.43)
(1000, 0.43)
(2000, 0.45)
(3000, 0.46)
(4000, 0.47)
(5000, 0.47)
(6000, 0.48)
(7000, 0.48)
(8000, 0.49)
(9000, 0.49)
(10000, 0.49)
(15000, 0.51)
(20000, 0.52)
(30000, 0.55)
(40000, 0.56)
(50000, 0.58)
(60000, 0.59)
};\addlegendentry{Perc train}
\addplot[
%dash pattern=on 4pt off 1pt on 4pt off 4pt,
%dashed,
color=blue,
%mark=*,
very thick,
error bars,
y dir= both,
y explicit
]
coordinates {
(20, 0.40)
(40, 0.43)
(60, 0.45)
(80, 0.46)
(100, 0.47)
(200, 0.50)
(300, 0.51)
(400, 0.53)
(500, 0.53)
(600, 0.54)
(700, 0.54)
(800, 0.55)
(900, 0.55)
(1000, 0.56)
(2000, 0.57)
(3000, 0.57)
(4000, 0.57)
(5000, 0.57)
(6000, 0.57)
(7000, 0.57)
(8000, 0.57)
(9000, 0.57)
(10000, 0.57)
(15000, 0.56)
(20000, 0.56)
(30000, 0.56)
(40000, 0.56)
(50000, 0.56)
(55000, 0.56)
(60000, 0.56)
};\addlegendentry{Lin SVC val}
\addplot[
%dash pattern=on 4pt off 1pt on 4pt off 4pt,
dashed,
color=blue,
%mark=*,
very thick,
error bars,
y dir= both,
y explicit
]
coordinates {
(20, 0.40)
(40, 0.43)
(60, 0.45)
(80, 0.46)
(100, 0.47)
(200, 0.50)
(300, 0.52)
(400, 0.53)
(500, 0.54)
(600, 0.55)
(700, 0.55)
(800, 0.55)
(900, 0.56)
(1000, 0.56)
(2000, 0.58)
(3000, 0.59)
(4000, 0.60)
(5000, 0.60)
(6000, 0.61)
(7000, 0.61)
(8000, 0.61)
(9000, 0.62)
(10000, 0.62)
(15000, 0.63)
(20000, 0.65)
(30000, 0.66)
(40000, 0.67)
(50000, 0.68)
(60000, 0.69)
};\addlegendentry{Lin SVC train}
\end{axis}
\end{tikzpicture}
}
\subfigure[Unigrams \& Bigrams]{
\begin{tikzpicture}[scale=0.8]
\begin{axis}[
legend style={legend pos=south east, font=\tiny},
xlabel={No. of Features},
ylabel={Accuracy},
xmin=-500, xmax=60000,
ymin=0.3, ymax=0.80,
xtick={0, 10000, 20000, 30000, 40000, 50000, 60000, 70000},
ytick={0.0, 0.1, 0.2, 0.3, 0.4, 0.5, 0.6, 0.7, 0.8, 0.9, 1.0},
ymajorgrids=true,
grid style=dashed
]
\addplot[
%dash pattern=on 4pt off 1pt on 4pt off 4pt,
%dashed,
color=red,
%mark=*,
very thick,
error bars,
y dir= both,
y explicit
]
coordinates {
(20, 0.43)
(40, 0.45)
(60, 0.46)
(80, 0.47)
(100, 0.48)
(200, 0.50)
(300, 0.52)
(400, 0.53)
(500, 0.53)
(600, 0.54)
(700, 0.55)
(800, 0.55)
(900, 0.57)
(1000, 0.59)
(2000, 0.60)
(3000, 0.62)
(4000, 0.63)
(5000, 0.63)
(6000, 0.635)
(7000, 0.64)
(8000, 0.64)
(9000, 0.64)
(10000, 0.64)
(15000, 0.64)
(20000, 0.64)
(30000, 0.64)
(40000, 0.64)
(50000, 0.64)
(55000, 0.64)
(60000, 0.64)
};\addlegendentry{Log Reg val}
\addplot[
%dash pattern=on 4pt off 1pt on 4pt off 4pt,
dashed,
color=red,
%mark=*,
very thick,
error bars,
y dir= both,
y explicit
]
coordinates {
(20, 0.43)
(40, 0.44)
(60, 0.45)
(80, 0.46)
(100, 0.47)
(200, 0.50)
(300, 0.52)
(400, 0.53)
(500, 0.54)
(600, 0.54)
(700, 0.55)
(800, 0.56)
(900, 0.58)
(1000, 0.60)
(2000, 0.62)
(3000, 0.63)
(4000, 0.64)
(5000, 0.67)
(6000, 0.68)
(7000, 0.69)
(8000, 0.70)
(9000, 0.71)
(10000, 0.71)
(15000, 0.72)
(20000, 0.72)
(30000, 0.72)
(40000, 0.73)
(50000, 0.73)
(60000, 0.73)
};\addlegendentry{Log Reg train}
\addplot[
%dash pattern=on 4pt off 1pt on 4pt off 4pt,
%dashed,
color=yellow,
%mark=*,
very thick,
error bars,
y dir= both,
y explicit
]
coordinates {
(20, 0.42)
(40, 0.43)
(60, 0.42)
(80, 0.42)
(100, 0.43)
(200, 0.47)
(300, 0.49)
(400, 0.50)
(500, 0.51)
(600, 0.53)
(700, 0.54)
(800, 0.54)
(900, 0.56)
(1000, 0.56)
(2000, 0.56)
(3000, 0.57)
(4000, 0.57)
(5000, 0.57)
(6000, 0.57)
(7000, 0.57)
(8000, 0.57)
(9000, 0.57)
(10000, 0.56)
(15000, 0.56)
(20000, 0.56)
(30000, 0.56)
(40000, 0.56)
(50000, 0.56)
(55000, 0.56)
(60000, 0.56)
};\addlegendentry{NB val}
\addplot[
%dash pattern=on 4pt off 1pt on 4pt off 4pt,
dashed,
color=yellow,
%mark=*,
very thick,
error bars,
y dir= both,
y explicit
]
coordinates {
(20, 0.43)
(40, 0.42)
(60, 0.42)
(80, 0.42)
(100, 0.43)
(200, 0.45)
(300, 0.47)
(400, 0.49)
(500, 0.51)
(600, 0.52)
(700, 0.53)
(800, 0.54)
(900, 0.55)
(1000, 0.56)
(2000, 0.57)
(3000, 0.58)
(4000, 0.58)
(5000, 0.58)
(6000, 0.58)
(7000, 0.59)
(8000, 0.59)
(9000, 0.59)
(10000, 0.59)
(15000, 0.60)
(20000, 0.60)
(30000, 0.60)
(40000, 0.61)
(50000, 0.61)
(60000, 0.61)
};\addlegendentry{NB train}
\addplot[
%dash pattern=on 4pt off 1pt on 4pt off 4pt,
%dashed,
color=green,
%mark=*,
very thick,
error bars,
y dir= both,
y explicit
]
coordinates {
(20, 0.35)
(40, 0.36)
(60, 0.37)
(80, 0.38)
(100, 0.38)
(200, 0.39)
(300, 0.40)
(400, 0.41)
(500, 0.42)
(600, 0.43)
(700, 0.44)
(800, 0.45)
(900, 0.46)
(1000, 0.47)
(2000, 0.47)
(3000, 0.47)
(4000, 0.48)
(5000, 0.48)
(6000, 0.48)
(7000, 0.48)
(8000, 0.49)
(9000, 0.49)
(10000, 0.49)
(15000, 0.49)
(20000, 0.49)
(30000, 0.49)
(40000, 0.49)
(50000, 0.49)
(55000, 0.49)
(60000, 0.49)
};\addlegendentry{Perc val}
\addplot[
%dash pattern=on 4pt off 1pt on 4pt off 4pt,
dashed,
color=green,
%mark=*,
very thick,
error bars,
y dir= both,
y explicit
]
coordinates {
(20, 0.36)
(40, 0.37)
(60, 0.38)
(80, 0.39)
(100, 0.40)
(200, 0.42)
(300, 0.44)
(400, 0.46)
(500, 0.48)
(600, 0.50)
(700, 0.51)
(800, 0.51)
(900, 0.52)
(1000, 0.53)
(2000, 0.54)
(3000, 0.55)
(4000, 0.56)
(5000, 0.56)
(6000, 0.58)
(7000, 0.58)
(8000, 0.60)
(9000, 0.61)
(10000, 0.62)
(15000, 0.63)
(20000, 0.64)
(30000, 0.65)
(40000, 0.66)
(50000, 0.67)
(60000, 0.68)
};\addlegendentry{Perc train}
\addplot[
%dash pattern=on 4pt off 1pt on 4pt off 4pt,
%dashed,
color=blue,
%mark=*,
very thick,
error bars,
y dir= both,
y explicit
]
coordinates {
(20, 0.44)
(40, 0.46)
(60, 0.47)
(80, 0.48)
(100, 0.49)
(200, 0.50)
(300, 0.51)
(400, 0.53)
(500, 0.55)
(600, 0.56)
(700, 0.57)
(800, 0.59)
(900, 0.60)
(1000, 0.61)
(2000, 0.61)
(3000, 0.62)
(4000, 0.62)
(5000, 0.62)
(6000, 0.62)
(7000, 0.62)
(8000, 0.62)
(9000, 0.62)
(10000, 0.63)
(15000, 0.63)
(20000, 0.63)
(30000, 0.63)
(40000, 0.63)
(50000, 0.63)
(55000, 0.63)
(60000, 0.63)
};\addlegendentry{Lin SVC val}
\addplot[
%dash pattern=on 4pt off 1pt on 4pt off 4pt,
dashed,
color=blue,
%mark=*,
very thick,
error bars,
y dir= both,
y explicit
]
coordinates {
(20, 0.45)
(40, 0.43)
(60, 0.45)
(80, 0.46)
(100, 0.47)
(200, 0.50)
(300, 0.52)
(400, 0.54)
(500, 0.56)
(600, 0.57)
(700, 0.58)
(800, 0.60)
(900, 0.62)
(1000, 0.63)
(2000, 0.64)
(3000, 0.65)
(4000, 0.66)
(5000, 0.67)
(6000, 0.68)
(7000, 0.69)
(8000, 0.70)
(9000, 0.71)
(10000, 0.72)
(15000, 0.73)
(20000, 0.74)
(30000, 0.75)
(40000, 0.76)
(50000, 0.77)
(60000, 0.77)
};\addlegendentry{Lin SVC train}
\end{axis}
\end{tikzpicture}
}
\caption{Accuracy plots for (a) Unigrams, and (b) Unigrams \& Bigrams}
\label{}
\end{figure*}
%%%%%%%%%%%%%%%%%%%%%%%%%%%%%%%%%%%%%%%%%%%%%%%%%%%%%%%%%%%%%%%%%%%%%%%%%%%%%%%%
%%%%%%%%%%%%%%%%%%%%%%%%%%%%%%%%%%%%%%%%%%%%%%%%%%%%%%%%%%%%%%%%%%%%%%%%%%%%%%%%
%%%%%%%%%%%%%%%%%%%%%%%%%%%%%%%%%%%%%%%%%%%%%%%%%%%%%%%%%%%%%%%%%%%%%%%%%%%%%%%%

\subsection{Unigrams, Bigrams \& Trigrams}

The results for this feature extraction method are almost exactly the same as those for the `Unigrams \& Bigrams' method (Figures 3(b) and 4(b)), and we omit the graphs due to brevity of space. The best RMSE and accuracy scores are 0.78 and 64\%, achieved by logistic regression.

Adding trigrams to the previous model does not help, because trigrams repeat rarely. It is unlikely that two different user would use the exact same 3-tuple to describe a restaurant. The TF-IDF weighting technique weights almost all the 3-tuples as very rare, therefore they are not very useful as features.

\subsection{Latent Semantic Indexing (LSI)}
Figure 5 shows the results of the experiments with LSI. Figure 5(a) is a plot of the 1000 highest singular values. We see that the graph starts leveling out at 200. This means that the top 200 topics in the reviews are the most important ones for training the model. Next, we evaluate each classifer on feature vectors of length up to 200; the performance of each classifer is shown in Figure 5(b) and (c), for the validation fold\footnote{We could not obtain the training-fold RMSE and accuracy results due to time constraints.}.

Figures 5(b) and (c) show some interesting patterns. Perceptrons perform the worst as usual, however we see two spikes in the performance around 170 and 200 features, where the RMSE and accuracy suddenly improve. It is not clear why this happens, but it suggests that we should consider more than 200 features to see if more of these spikes occur and improve the best scores for perceptrons. The plots for logistic regression show another interesting pattern. As the number of features increases, the RMSE remains constant around 1.3, but the accuracy increases from 0.34 to 0.43. One explanation for this apparent anomaly is that, more examples get classified accurately, but the RMSE for the misclassified examples increases and overshadows the decrease in the RMSE due to better accuracy. 

A similar pattern is seen for Linear SVC, and could be explained as above. The upward trend in its accuracy suggests we should consider more than 200 features. Due to time constraints, we leave this experiment as future work.  
%%%%%%%%%%%%%%%%%%%%%%%%%%%%%%%%%%%%%%%%%%%%%%%%%%%%%%%%%%%%%%%%%%%%%%%%%%%%%%%%%%%%%%%%%%%%%%%%%%%%
%%%%%%%%%%%%%%%%%%%%%%%%%%%%%%%%%%%%%%%%%%%%%%%%%%%%%%%%%%%%%%%%%%%%%%%%%%%%%%%%%%%%%%%%%%%%%%%%%%%%
%%%%%%%%%%%%%%%%%%%%%%%%%%%%%%%%%%%%%%%%%%%%%%%%%%%%%%%%%%%%%%%%%%%%%%%%%%%%%%%%%%%%%%%%%%%%%%%%%%%%

\begin{figure}[ht!]
\centering
\subfigure[]{
\begin{tikzpicture}[scale=0.8]
\begin{axis}[
xlabel={No. of Singular Values},
ylabel={Singular Values},
xmin=-30, xmax=1000,
ymin=0, ymax=130,
xtick={0, 200, 400,600,800,1000},
ytick={0.0, 20, 40, 60, 80, 100, 120},
ymajorgrids=true,
grid style=dashed
]
\addplot[
%dash pattern=on 4pt off 1pt on 4pt off 4pt,
%dashed,
color=blue,
mark=*,
only marks,
mark options={fill=white},
%very thick,
error bars,
y dir= both,
y explicit
]
coordinates {
(	1000	,	10.4807952	)
(	999	,	10.48591001	)
(	998	,	10.49134862	)
(	997	,	10.49404423	)
(	996	,	10.50612312	)
(	995	,	10.51035004	)
(	994	,	10.51529226	)
(	993	,	10.52003966	)
(	992	,	10.52330276	)
(	991	,	10.52762289	)
(	990	,	10.5341564	)
(	989	,	10.54422748	)
(	988	,	10.54855784	)
(	987	,	10.55265763	)
(	986	,	10.55662207	)
(	985	,	10.55884808	)
(	984	,	10.56398613	)
(	983	,	10.57185354	)
(	982	,	10.57502804	)
(	981	,	10.57887283	)
(	980	,	10.58600395	)
(	979	,	10.59287641	)
(	978	,	10.59979102	)
(	977	,	10.6046544	)
(	976	,	10.60935163	)
(	975	,	10.61788348	)
(	974	,	10.62316879	)
(	973	,	10.62578496	)
(	972	,	10.6293134	)
(	971	,	10.63338468	)
(	970	,	10.63513654	)
(	969	,	10.63600035	)
(	968	,	10.64274643	)
(	967	,	10.6465551	)
(	966	,	10.65214783	)
(	965	,	10.65905633	)
(	964	,	10.6608467	)
(	963	,	10.66292313	)
(	962	,	10.66660322	)
(	961	,	10.67532262	)
(	960	,	10.68146378	)
(	959	,	10.68522571	)
(	958	,	10.68687062	)
(	957	,	10.69649963	)
(	956	,	10.7036957	)
(	955	,	10.70495737	)
(	954	,	10.70544043	)
(	953	,	10.71120277	)
(	952	,	10.7153117	)
(	951	,	10.72642358	)
(	950	,	10.72950957	)
(	949	,	10.73830909	)
(	948	,	10.74225774	)
(	947	,	10.74521629	)
(	946	,	10.74981715	)
(	945	,	10.75679452	)
(	944	,	10.75961968	)
(	943	,	10.76723019	)
(	942	,	10.77042774	)
(	941	,	10.77253411	)
(	940	,	10.77624277	)
(	939	,	10.77913762	)
(	938	,	10.78360207	)
(	937	,	10.78720795	)
(	936	,	10.78941924	)
(	935	,	10.79730564	)
(	934	,	10.80210635	)
(	933	,	10.80631315	)
(	932	,	10.81142958	)
(	931	,	10.81471012	)
(	930	,	10.81995973	)
(	929	,	10.8265852	)
(	928	,	10.83328634	)
(	927	,	10.83642065	)
(	926	,	10.83907917	)
(	925	,	10.84274897	)
(	924	,	10.84620733	)
(	923	,	10.84992117	)
(	922	,	10.85502703	)
(	921	,	10.85826297	)
(	920	,	10.86043881	)
(	919	,	10.86849568	)
(	918	,	10.87088657	)
(	917	,	10.87684238	)
(	916	,	10.88071118	)
(	915	,	10.88344713	)
(	914	,	10.88820956	)
(	913	,	10.89227712	)
(	912	,	10.89598908	)
(	911	,	10.89962725	)
(	910	,	10.90754331	)
(	909	,	10.90886344	)
(	908	,	10.91451875	)
(	907	,	10.91963438	)
(	906	,	10.92466361	)
(	905	,	10.92696432	)
(	904	,	10.93385673	)
(	903	,	10.93970332	)
(	902	,	10.94459782	)
(	901	,	10.9474986	)
(	900	,	10.951038	)
(	899	,	10.95490642	)
(	898	,	10.95813903	)
(	897	,	10.96478661	)
(	896	,	10.96869433	)
(	895	,	10.97235443	)
(	894	,	10.97987413	)
(	893	,	10.98451267	)
(	892	,	10.98598261	)
(	891	,	10.99210609	)
(	890	,	10.99631332	)
(	889	,	10.99999055	)
(	888	,	11.00614908	)
(	887	,	11.00813347	)
(	886	,	11.01110395	)
(	885	,	11.0190395	)
(	884	,	11.02034746	)
(	883	,	11.02970948	)
(	882	,	11.03363543	)
(	881	,	11.03637068	)
(	880	,	11.03949911	)
(	879	,	11.04266536	)
(	878	,	11.04692759	)
(	877	,	11.04960183	)
(	876	,	11.05416406	)
(	875	,	11.05912055	)
(	874	,	11.06549219	)
(	873	,	11.0703291	)
(	872	,	11.07785644	)
(	871	,	11.08344706	)
(	870	,	11.08708741	)
(	869	,	11.08778865	)
(	868	,	11.09166994	)
(	867	,	11.09757511	)
(	866	,	11.09997363	)
(	865	,	11.11109921	)
(	864	,	11.11486554	)
(	863	,	11.12088631	)
(	862	,	11.12433914	)
(	861	,	11.12480769	)
(	860	,	11.13030092	)
(	859	,	11.14024122	)
(	858	,	11.14598928	)
(	857	,	11.14933329	)
(	856	,	11.15031722	)
(	855	,	11.15612248	)
(	854	,	11.15780426	)
(	853	,	11.16625246	)
(	852	,	11.16820178	)
(	851	,	11.1693444	)
(	850	,	11.17361365	)
(	849	,	11.18088234	)
(	848	,	11.18303297	)
(	847	,	11.18993667	)
(	846	,	11.19612425	)
(	845	,	11.19956067	)
(	844	,	11.20366344	)
(	843	,	11.20416987	)
(	842	,	11.21057759	)
(	841	,	11.21530036	)
(	840	,	11.21843855	)
(	839	,	11.22331742	)
(	838	,	11.22915088	)
(	837	,	11.23219831	)
(	836	,	11.23670093	)
(	835	,	11.24523341	)
(	834	,	11.25357398	)
(	833	,	11.25834862	)
(	832	,	11.25922808	)
(	831	,	11.26688871	)
(	830	,	11.26844604	)
(	829	,	11.2704294	)
(	828	,	11.27717032	)
(	827	,	11.28184595	)
(	826	,	11.28390319	)
(	825	,	11.28741265	)
(	824	,	11.29796331	)
(	823	,	11.30592897	)
(	822	,	11.30875022	)
(	821	,	11.31111637	)
(	820	,	11.32073312	)
(	819	,	11.32348412	)
(	818	,	11.32882683	)
(	817	,	11.33539306	)
(	816	,	11.33989635	)
(	815	,	11.34448186	)
(	814	,	11.34704387	)
(	813	,	11.34963141	)
(	812	,	11.35364712	)
(	811	,	11.35533131	)
(	810	,	11.36701471	)
(	809	,	11.37985241	)
(	808	,	11.38514863	)
(	807	,	11.39164134	)
(	806	,	11.39510327	)
(	805	,	11.39993379	)
(	804	,	11.40359886	)
(	803	,	11.40710363	)
(	802	,	11.41294064	)
(	801	,	11.42315825	)
(	800	,	11.42700455	)
(	799	,	11.43471769	)
(	798	,	11.43753131	)
(	797	,	11.44488125	)
(	796	,	11.4483131	)
(	795	,	11.45508073	)
(	794	,	11.45955625	)
(	793	,	11.46798255	)
(	792	,	11.47913768	)
(	791	,	11.48192695	)
(	790	,	11.48863513	)
(	789	,	11.49122644	)
(	788	,	11.49939685	)
(	787	,	11.50206842	)
(	786	,	11.50772828	)
(	785	,	11.51426347	)
(	784	,	11.52268944	)
(	783	,	11.52492286	)
(	782	,	11.53967925	)
(	781	,	11.55053829	)
(	780	,	11.55306479	)
(	779	,	11.56125362	)
(	778	,	11.56603058	)
(	777	,	11.56645817	)
(	776	,	11.57201521	)
(	775	,	11.58606266	)
(	774	,	11.58887315	)
(	773	,	11.59255726	)
(	772	,	11.59733075	)
(	771	,	11.60083091	)
(	770	,	11.60406141	)
(	769	,	11.62083508	)
(	768	,	11.62475904	)
(	767	,	11.63491911	)
(	766	,	11.64393119	)
(	765	,	11.64602525	)
(	764	,	11.65166669	)
(	763	,	11.66824793	)
(	762	,	11.67225333	)
(	761	,	11.67834709	)
(	760	,	11.68595289	)
(	759	,	11.69526839	)
(	758	,	11.69952997	)
(	757	,	11.70791787	)
(	756	,	11.71143894	)
(	755	,	11.71890471	)
(	754	,	11.73279276	)
(	753	,	11.73654245	)
(	752	,	11.74308644	)
(	751	,	11.74749578	)
(	750	,	11.75189007	)
(	749	,	11.76234124	)
(	748	,	11.76755996	)
(	747	,	11.77258685	)
(	746	,	11.78120209	)
(	745	,	11.78664575	)
(	744	,	11.79390549	)
(	743	,	11.80563112	)
(	742	,	11.80827609	)
(	741	,	11.82388627	)
(	740	,	11.82596012	)
(	739	,	11.8331097	)
(	738	,	11.84525463	)
(	737	,	11.84785303	)
(	736	,	11.85485959	)
(	735	,	11.86087955	)
(	734	,	11.8700762	)
(	733	,	11.87968409	)
(	732	,	11.88254369	)
(	731	,	11.89986717	)
(	730	,	11.90218686	)
(	729	,	11.90821628	)
(	728	,	11.90858701	)
(	727	,	11.91188851	)
(	726	,	11.92192738	)
(	725	,	11.92877232	)
(	724	,	11.93529157	)
(	723	,	11.95090896	)
(	722	,	11.95348355	)
(	721	,	11.9654547	)
(	720	,	11.97196485	)
(	719	,	11.98066613	)
(	718	,	11.9856852	)
(	717	,	11.99270116	)
(	716	,	11.99829867	)
(	715	,	12.01091459	)
(	714	,	12.0129951	)
(	713	,	12.01786409	)
(	712	,	12.02642758	)
(	711	,	12.03353262	)
(	710	,	12.03835632	)
(	709	,	12.05561449	)
(	708	,	12.06460708	)
(	707	,	12.07360494	)
(	706	,	12.07801548	)
(	705	,	12.08296953	)
(	704	,	12.09426944	)
(	703	,	12.09878466	)
(	702	,	12.10284703	)
(	701	,	12.11180283	)
(	700	,	12.11297915	)
(	699	,	12.12216761	)
(	698	,	12.12911755	)
(	697	,	12.13115045	)
(	696	,	12.1357526	)
(	695	,	12.15208701	)
(	694	,	12.15501874	)
(	693	,	12.16142461	)
(	692	,	12.16600605	)
(	691	,	12.17856285	)
(	690	,	12.17921207	)
(	689	,	12.18991849	)
(	688	,	12.19699708	)
(	687	,	12.20616583	)
(	686	,	12.21531354	)
(	685	,	12.22279883	)
(	684	,	12.24147278	)
(	683	,	12.24515067	)
(	682	,	12.25058733	)
(	681	,	12.25712084	)
(	680	,	12.26339691	)
(	679	,	12.27148106	)
(	678	,	12.28267373	)
(	677	,	12.28548718	)
(	676	,	12.29000484	)
(	675	,	12.30303561	)
(	674	,	12.30788182	)
(	673	,	12.32101635	)
(	672	,	12.32803499	)
(	671	,	12.33034693	)
(	670	,	12.34209015	)
(	669	,	12.34932727	)
(	668	,	12.35229038	)
(	667	,	12.36218315	)
(	666	,	12.37205468	)
(	665	,	12.37551075	)
(	664	,	12.38364666	)
(	663	,	12.39224323	)
(	662	,	12.40214241	)
(	661	,	12.40500597	)
(	660	,	12.4194435	)
(	659	,	12.42509017	)
(	658	,	12.42831634	)
(	657	,	12.43714798	)
(	656	,	12.44901125	)
(	655	,	12.45470475	)
(	654	,	12.46061054	)
(	653	,	12.46446671	)
(	652	,	12.47081634	)
(	651	,	12.49353085	)
(	650	,	12.50086513	)
(	649	,	12.51187496	)
(	648	,	12.51469351	)
(	647	,	12.52546554	)
(	646	,	12.53334358	)
(	645	,	12.54256546	)
(	644	,	12.55652515	)
(	643	,	12.5669865	)
(	642	,	12.57849742	)
(	641	,	12.58438472	)
(	640	,	12.58633634	)
(	639	,	12.59749535	)
(	638	,	12.60173192	)
(	637	,	12.60606685	)
(	636	,	12.61377996	)
(	635	,	12.62583925	)
(	634	,	12.63627809	)
(	633	,	12.6512018	)
(	632	,	12.65762859	)
(	631	,	12.6716475	)
(	630	,	12.67560712	)
(	629	,	12.68273134	)
(	628	,	12.69096647	)
(	627	,	12.69479211	)
(	626	,	12.71152903	)
(	625	,	12.71489107	)
(	624	,	12.73454413	)
(	623	,	12.73718825	)
(	622	,	12.74804776	)
(	621	,	12.75134384	)
(	620	,	12.75543886	)
(	619	,	12.76779444	)
(	618	,	12.77700584	)
(	617	,	12.78014564	)
(	616	,	12.79516497	)
(	615	,	12.80692149	)
(	614	,	12.81253548	)
(	613	,	12.81686838	)
(	612	,	12.8258508	)
(	611	,	12.83383511	)
(	610	,	12.84387245	)
(	609	,	12.84991344	)
(	608	,	12.85552427	)
(	607	,	12.86105046	)
(	606	,	12.86680538	)
(	605	,	12.87495115	)
(	604	,	12.88627154	)
(	603	,	12.88996164	)
(	602	,	12.89589705	)
(	601	,	12.90643997	)
(	600	,	12.91010425	)
(	599	,	12.91959423	)
(	598	,	12.92944031	)
(	597	,	12.93768253	)
(	596	,	12.94738862	)
(	595	,	12.95863668	)
(	594	,	12.96266405	)
(	593	,	12.9671936	)
(	592	,	12.97858655	)
(	591	,	12.98266179	)
(	590	,	12.99431844	)
(	589	,	12.99719271	)
(	588	,	13.00320428	)
(	587	,	13.00749704	)
(	586	,	13.0100816	)
(	585	,	13.02769906	)
(	584	,	13.03455617	)
(	583	,	13.04694931	)
(	582	,	13.05162908	)
(	581	,	13.05450657	)
(	580	,	13.06611329	)
(	579	,	13.07070786	)
(	578	,	13.07482074	)
(	577	,	13.08936435	)
(	576	,	13.09382782	)
(	575	,	13.11158653	)
(	574	,	13.11380594	)
(	573	,	13.12698722	)
(	572	,	13.13082126	)
(	571	,	13.13851202	)
(	570	,	13.14492738	)
(	569	,	13.14603208	)
(	568	,	13.16245174	)
(	567	,	13.16920666	)
(	566	,	13.17466939	)
(	565	,	13.17880178	)
(	564	,	13.18466155	)
(	563	,	13.19169013	)
(	562	,	13.20072063	)
(	561	,	13.2087888	)
(	560	,	13.21759768	)
(	559	,	13.22793809	)
(	558	,	13.23198026	)
(	557	,	13.24759774	)
(	556	,	13.26682292	)
(	555	,	13.28078311	)
(	554	,	13.28252918	)
(	553	,	13.29415128	)
(	552	,	13.3076231	)
(	551	,	13.31591115	)
(	550	,	13.32345432	)
(	549	,	13.32717998	)
(	548	,	13.343091	)
(	547	,	13.3601549	)
(	546	,	13.37099793	)
(	545	,	13.37692479	)
(	544	,	13.38664842	)
(	543	,	13.3992148	)
(	542	,	13.40530041	)
(	541	,	13.41442379	)
(	540	,	13.43006175	)
(	539	,	13.44880296	)
(	538	,	13.45400496	)
(	537	,	13.45683627	)
(	536	,	13.47067552	)
(	535	,	13.48340236	)
(	534	,	13.48999043	)
(	533	,	13.49449406	)
(	532	,	13.51393306	)
(	531	,	13.51875821	)
(	530	,	13.54596191	)
(	529	,	13.55088058	)
(	528	,	13.56114581	)
(	527	,	13.56879284	)
(	526	,	13.57502793	)
(	525	,	13.59363792	)
(	524	,	13.59943625	)
(	523	,	13.60868123	)
(	522	,	13.6100665	)
(	521	,	13.62100911	)
(	520	,	13.63336782	)
(	519	,	13.64498669	)
(	518	,	13.65365731	)
(	517	,	13.66445872	)
(	516	,	13.67197809	)
(	515	,	13.68091412	)
(	514	,	13.68470376	)
(	513	,	13.7044076	)
(	512	,	13.7169676	)
(	511	,	13.72823669	)
(	510	,	13.72938345	)
(	509	,	13.75288904	)
(	508	,	13.75418457	)
(	507	,	13.76431258	)
(	506	,	13.77435089	)
(	505	,	13.77707767	)
(	504	,	13.79598272	)
(	503	,	13.80877758	)
(	502	,	13.81699512	)
(	501	,	13.82899566	)
(	500	,	13.83664955	)
(	499	,	13.84402749	)
(	498	,	13.8587083	)
(	497	,	13.86462802	)
(	496	,	13.8785394	)
(	495	,	13.88732305	)
(	494	,	13.89212978	)
(	493	,	13.89373728	)
(	492	,	13.89768244	)
(	491	,	13.90432178	)
(	490	,	13.91148662	)
(	489	,	13.93300544	)
(	488	,	13.94210908	)
(	487	,	13.95303618	)
(	486	,	13.96599911	)
(	485	,	13.96943052	)
(	484	,	13.99343073	)
(	483	,	13.99757677	)
(	482	,	14.00104699	)
(	481	,	14.01124192	)
(	480	,	14.02292271	)
(	479	,	14.03161018	)
(	478	,	14.03796441	)
(	477	,	14.04228855	)
(	476	,	14.05112531	)
(	475	,	14.05900718	)
(	474	,	14.08324419	)
(	473	,	14.09372203	)
(	472	,	14.09474823	)
(	471	,	14.12029073	)
(	470	,	14.1241375	)
(	469	,	14.13302473	)
(	468	,	14.14059379	)
(	467	,	14.15438501	)
(	466	,	14.17633486	)
(	465	,	14.18921645	)
(	464	,	14.20168398	)
(	463	,	14.21441824	)
(	462	,	14.22786997	)
(	461	,	14.23268469	)
(	460	,	14.24737867	)
(	459	,	14.25427477	)
(	458	,	14.25867911	)
(	457	,	14.27018193	)
(	456	,	14.27577529	)
(	455	,	14.28785943	)
(	454	,	14.29661906	)
(	453	,	14.30449365	)
(	452	,	14.30962738	)
(	451	,	14.3272606	)
(	450	,	14.3317902	)
(	449	,	14.34419026	)
(	448	,	14.35601179	)
(	447	,	14.36053421	)
(	446	,	14.37957884	)
(	445	,	14.388204	)
(	444	,	14.4025578	)
(	443	,	14.42234211	)
(	442	,	14.42774948	)
(	441	,	14.43645032	)
(	440	,	14.44642609	)
(	439	,	14.45834022	)
(	438	,	14.46363762	)
(	437	,	14.47701063	)
(	436	,	14.48302909	)
(	435	,	14.49705457	)
(	434	,	14.50237311	)
(	433	,	14.52083486	)
(	432	,	14.52603836	)
(	431	,	14.53112626	)
(	430	,	14.54465061	)
(	429	,	14.54765637	)
(	428	,	14.5725531	)
(	427	,	14.58366553	)
(	426	,	14.59249861	)
(	425	,	14.60533455	)
(	424	,	14.61078733	)
(	423	,	14.62854207	)
(	422	,	14.64084042	)
(	421	,	14.65295369	)
(	420	,	14.66091533	)
(	419	,	14.67836778	)
(	418	,	14.68739246	)
(	417	,	14.69902304	)
(	416	,	14.70410551	)
(	415	,	14.71985444	)
(	414	,	14.74369058	)
(	413	,	14.75459734	)
(	412	,	14.76290114	)
(	411	,	14.77109282	)
(	410	,	14.7865354	)
(	409	,	14.8091712	)
(	408	,	14.82288849	)
(	407	,	14.82701574	)
(	406	,	14.84796891	)
(	405	,	14.84979005	)
(	404	,	14.86187756	)
(	403	,	14.86710256	)
(	402	,	14.87415909	)
(	401	,	14.90444354	)
(	400	,	14.91773172	)
(	399	,	14.92649201	)
(	398	,	14.93696328	)
(	397	,	14.94875102	)
(	396	,	14.9673156	)
(	395	,	14.97289884	)
(	394	,	14.98997959	)
(	393	,	14.99593778	)
(	392	,	15.01361593	)
(	391	,	15.0269158	)
(	390	,	15.05369486	)
(	389	,	15.0659016	)
(	388	,	15.06791942	)
(	387	,	15.08226491	)
(	386	,	15.09127107	)
(	385	,	15.11132131	)
(	384	,	15.12803451	)
(	383	,	15.13615815	)
(	382	,	15.14704225	)
(	381	,	15.15722377	)
(	380	,	15.17606599	)
(	379	,	15.19374713	)
(	378	,	15.20339082	)
(	377	,	15.21495524	)
(	376	,	15.2312222	)
(	375	,	15.24341734	)
(	374	,	15.25006697	)
(	373	,	15.2718144	)
(	372	,	15.27709383	)
(	371	,	15.29549516	)
(	370	,	15.30081497	)
(	369	,	15.3073445	)
(	368	,	15.32866951	)
(	367	,	15.34015241	)
(	366	,	15.35705627	)
(	365	,	15.37187425	)
(	364	,	15.38484311	)
(	363	,	15.39394976	)
(	362	,	15.4146293	)
(	361	,	15.42925273	)
(	360	,	15.4423552	)
(	359	,	15.4474926	)
(	358	,	15.46278873	)
(	357	,	15.48021486	)
(	356	,	15.49649971	)
(	355	,	15.50786651	)
(	354	,	15.5183766	)
(	353	,	15.52925281	)
(	352	,	15.54392697	)
(	351	,	15.55754384	)
(	350	,	15.56947582	)
(	349	,	15.61067538	)
(	348	,	15.61565583	)
(	347	,	15.6361631	)
(	346	,	15.64628217	)
(	345	,	15.66527131	)
(	344	,	15.68031309	)
(	343	,	15.70084452	)
(	342	,	15.70876508	)
(	341	,	15.71733231	)
(	340	,	15.75402679	)
(	339	,	15.77108985	)
(	338	,	15.78867413	)
(	337	,	15.79582942	)
(	336	,	15.8305769	)
(	335	,	15.83835436	)
(	334	,	15.86954092	)
(	333	,	15.8943426	)
(	332	,	15.91015346	)
(	331	,	15.91596994	)
(	330	,	15.95186166	)
(	329	,	15.96554006	)
(	328	,	15.98033607	)
(	327	,	15.9902937	)
(	326	,	16.03174142	)
(	325	,	16.05506492	)
(	324	,	16.06315962	)
(	323	,	16.0809386	)
(	322	,	16.11125147	)
(	321	,	16.11953023	)
(	320	,	16.13789462	)
(	319	,	16.15728803	)
(	318	,	16.17066157	)
(	317	,	16.19923071	)
(	316	,	16.2126367	)
(	315	,	16.22793614	)
(	314	,	16.23457308	)
(	313	,	16.24570654	)
(	312	,	16.25542539	)
(	311	,	16.27506275	)
(	310	,	16.29328496	)
(	309	,	16.31568759	)
(	308	,	16.33218263	)
(	307	,	16.35313962	)
(	306	,	16.35461398	)
(	305	,	16.40003544	)
(	304	,	16.42869325	)
(	303	,	16.434453	)
(	302	,	16.44613606	)
(	301	,	16.45696837	)
(	300	,	16.46581113	)
(	299	,	16.47039256	)
(	298	,	16.48837154	)
(	297	,	16.51148405	)
(	296	,	16.52669878	)
(	295	,	16.53889766	)
(	294	,	16.56333078	)
(	293	,	16.57180711	)
(	292	,	16.5875544	)
(	291	,	16.59866091	)
(	290	,	16.60612182	)
(	289	,	16.62141409	)
(	288	,	16.64583424	)
(	287	,	16.65995112	)
(	286	,	16.66642483	)
(	285	,	16.70010426	)
(	284	,	16.72983811	)
(	283	,	16.74980463	)
(	282	,	16.77101478	)
(	281	,	16.78530617	)
(	280	,	16.80137474	)
(	279	,	16.82326717	)
(	278	,	16.8446744	)
(	277	,	16.85456459	)
(	276	,	16.88193133	)
(	275	,	16.90080419	)
(	274	,	16.91097989	)
(	273	,	16.93984796	)
(	272	,	16.94984805	)
(	271	,	16.95834111	)
(	270	,	16.99116202	)
(	269	,	17.02054261	)
(	268	,	17.03742713	)
(	267	,	17.07402072	)
(	266	,	17.07622402	)
(	265	,	17.09419641	)
(	264	,	17.10908526	)
(	263	,	17.13973125	)
(	262	,	17.14276038	)
(	261	,	17.15123413	)
(	260	,	17.17819347	)
(	259	,	17.20572619	)
(	258	,	17.21528907	)
(	257	,	17.23232565	)
(	256	,	17.24985585	)
(	255	,	17.26695813	)
(	254	,	17.29748585	)
(	253	,	17.32155385	)
(	252	,	17.33469331	)
(	251	,	17.34821018	)
(	250	,	17.40866961	)
(	249	,	17.42791246	)
(	248	,	17.43256725	)
(	247	,	17.46554442	)
(	246	,	17.47252527	)
(	245	,	17.50825599	)
(	244	,	17.53182254	)
(	243	,	17.56041677	)
(	242	,	17.58333547	)
(	241	,	17.61064347	)
(	240	,	17.65589283	)
(	239	,	17.68285235	)
(	238	,	17.70186631	)
(	237	,	17.73300669	)
(	236	,	17.73441828	)
(	235	,	17.77025212	)
(	234	,	17.81385946	)
(	233	,	17.8497071	)
(	232	,	17.86537115	)
(	231	,	17.87929522	)
(	230	,	17.89552653	)
(	229	,	17.91113297	)
(	228	,	17.93441314	)
(	227	,	17.95633991	)
(	226	,	17.98534407	)
(	225	,	18.01598782	)
(	224	,	18.028966	)
(	223	,	18.09996597	)
(	222	,	18.1161963	)
(	221	,	18.13152001	)
(	220	,	18.1594548	)
(	219	,	18.20385473	)
(	218	,	18.22489543	)
(	217	,	18.22909056	)
(	216	,	18.30384316	)
(	215	,	18.33183611	)
(	214	,	18.34205516	)
(	213	,	18.36058444	)
(	212	,	18.38439697	)
(	211	,	18.41154009	)
(	210	,	18.43081967	)
(	209	,	18.46637492	)
(	208	,	18.47635643	)
(	207	,	18.49361908	)
(	206	,	18.53812162	)
(	205	,	18.55649409	)
(	204	,	18.57141762	)
(	203	,	18.60102023	)
(	202	,	18.62969618	)
(	201	,	18.64012693	)
(	200	,	18.66915658	)
(	199	,	18.73833125	)
(	198	,	18.7672177	)
(	197	,	18.78105575	)
(	196	,	18.80649339	)
(	195	,	18.91301083	)
(	194	,	18.91481171	)
(	193	,	18.94840534	)
(	192	,	18.95188926	)
(	191	,	19.01169499	)
(	190	,	19.04633501	)
(	189	,	19.05650394	)
(	188	,	19.10269891	)
(	187	,	19.16357555	)
(	186	,	19.17082875	)
(	185	,	19.17508068	)
(	184	,	19.24717483	)
(	183	,	19.28366787	)
(	182	,	19.30946748	)
(	181	,	19.31979361	)
(	180	,	19.39671862	)
(	179	,	19.41779291	)
(	178	,	19.4606897	)
(	177	,	19.51376237	)
(	176	,	19.54922311	)
(	175	,	19.61703095	)
(	174	,	19.62547891	)
(	173	,	19.67382273	)
(	172	,	19.68721419	)
(	171	,	19.7242266	)
(	170	,	19.77231474	)
(	169	,	19.78484816	)
(	168	,	19.85338266	)
(	167	,	19.88159462	)
(	166	,	19.89675003	)
(	165	,	19.94894793	)
(	164	,	19.98125593	)
(	163	,	19.99503169	)
(	162	,	20.01881751	)
(	161	,	20.07033098	)
(	160	,	20.07805584	)
(	159	,	20.12294215	)
(	158	,	20.14685441	)
(	157	,	20.18144192	)
(	156	,	20.22752961	)
(	155	,	20.25530336	)
(	154	,	20.28321717	)
(	153	,	20.31777325	)
(	152	,	20.32451997	)
(	151	,	20.36377208	)
(	150	,	20.40100353	)
(	149	,	20.42210379	)
(	148	,	20.4470955	)
(	147	,	20.46343986	)
(	146	,	20.51764885	)
(	145	,	20.53064294	)
(	144	,	20.54667927	)
(	143	,	20.58751236	)
(	142	,	20.62083369	)
(	141	,	20.65092948	)
(	140	,	20.72178218	)
(	139	,	20.76368544	)
(	138	,	20.78345666	)
(	137	,	20.81504766	)
(	136	,	20.83679732	)
(	135	,	20.9069332	)
(	134	,	20.93167221	)
(	133	,	20.94901917	)
(	132	,	21.00991194	)
(	131	,	21.03872142	)
(	130	,	21.07364238	)
(	129	,	21.0892989	)
(	128	,	21.1197189	)
(	127	,	21.14278867	)
(	126	,	21.20492682	)
(	125	,	21.23766312	)
(	124	,	21.25140733	)
(	123	,	21.29715251	)
(	122	,	21.30917317	)
(	121	,	21.38771036	)
(	120	,	21.47799879	)
(	119	,	21.48979519	)
(	118	,	21.52023921	)
(	117	,	21.56915488	)
(	116	,	21.61518771	)
(	115	,	21.63525532	)
(	114	,	21.65862351	)
(	113	,	21.68276188	)
(	112	,	21.72338274	)
(	111	,	21.77774425	)
(	110	,	21.84379267	)
(	109	,	21.85876125	)
(	108	,	21.87087976	)
(	107	,	21.90601798	)
(	106	,	21.97740363	)
(	105	,	22.04291341	)
(	104	,	22.1005755	)
(	103	,	22.12705896	)
(	102	,	22.14217664	)
(	101	,	22.19925657	)
(	100	,	22.30153077	)
(	99	,	22.34435594	)
(	98	,	22.36772963	)
(	97	,	22.51198682	)
(	96	,	22.56056585	)
(	95	,	22.57182289	)
(	94	,	22.65366987	)
(	93	,	22.71761799	)
(	92	,	22.74328071	)
(	91	,	22.84499213	)
(	90	,	22.8681606	)
(	89	,	22.97673833	)
(	88	,	23.04450842	)
(	87	,	23.14499241	)
(	86	,	23.15875804	)
(	85	,	23.21908232	)
(	84	,	23.32402373	)
(	83	,	23.36660663	)
(	82	,	23.40268219	)
(	81	,	23.47915587	)
(	80	,	23.60903996	)
(	79	,	23.63575597	)
(	78	,	23.80623077	)
(	77	,	23.89642805	)
(	76	,	23.92956291	)
(	75	,	24.0287033	)
(	74	,	24.11526167	)
(	73	,	24.20788629	)
(	72	,	24.31391403	)
(	71	,	24.35670566	)
(	70	,	24.49164152	)
(	69	,	24.51018723	)
(	68	,	24.57373715	)
(	67	,	24.62998826	)
(	66	,	24.80694558	)
(	65	,	24.82151547	)
(	64	,	25.01902059	)
(	63	,	25.06626902	)
(	62	,	25.09354618	)
(	61	,	25.27599477	)
(	60	,	25.2989506	)
(	59	,	25.45358059	)
(	58	,	25.46438512	)
(	57	,	25.58881892	)
(	56	,	25.63872859	)
(	55	,	25.75948108	)
(	54	,	25.78375661	)
(	53	,	25.87313628	)
(	52	,	25.96284254	)
(	51	,	26.09212574	)
(	50	,	26.26524218	)
(	49	,	26.35269872	)
(	48	,	26.41374768	)
(	47	,	26.53001431	)
(	46	,	26.67083214	)
(	45	,	26.85846672	)
(	44	,	26.9078818	)
(	43	,	27.11668668	)
(	42	,	27.21313086	)
(	41	,	27.40033803	)
(	40	,	27.48038007	)
(	39	,	27.55186478	)
(	38	,	27.70499641	)
(	37	,	27.99674008	)
(	36	,	28.3081872	)
(	35	,	28.43900551	)
(	34	,	28.56178499	)
(	33	,	28.81802217	)
(	32	,	28.99763069	)
(	31	,	29.0108095	)
(	30	,	29.35704433	)
(	29	,	30.11189248	)
(	28	,	30.22404098	)
(	27	,	30.36444662	)
(	26	,	30.53576948	)
(	25	,	30.71552324	)
(	24	,	31.06030211	)
(	23	,	31.39009469	)
(	22	,	31.6457979	)
(	21	,	32.00506804	)
(	20	,	32.53677638	)
(	19	,	32.64757329	)
(	18	,	33.16141976	)
(	17	,	33.88256874	)
(	16	,	34.56618953	)
(	15	,	35.14759888	)
(	14	,	35.57747406	)
(	13	,	36.02644833	)
(	12	,	36.73343342	)
(	11	,	37.25402285	)
(	10	,	38.70562062	)
(	9	,	40.17630127	)
(	8	,	41.05771638	)
(	7	,	41.87796446	)
(	6	,	43.21953101	)
(	5	,	43.86940184	)
(	4	,	44.69842133	)
(	3	,	50.39325134	)
(	2	,	54.15604511	)
(	1	,	122.91471133	)
};%\addlegendentry{Log Reg val}
\end{axis}
\end{tikzpicture}
}\\
\subfigure[]{
\begin{tikzpicture}[scale=0.8]
\begin{axis}[
legend style={legend pos=north east, font=\tiny},
xlabel={No. of Features},
ylabel={RMSE},
xmin=-1, xmax=202,
ymin=1.2, ymax=2.4,
xtick={0,20, 40,60, 80,100, 120,140, 160,180, 200},
ytick={1.2, 1.4, 1.6, 1.8, 2.0, 2.2, 2.4},
ymajorgrids=true,
grid style=dashed
]
\addplot[
%dash pattern=on 4pt off 1pt on 4pt off 4pt,
%dashed,
color=red,
%mark=*,
very thick,
error bars,
y dir= both,
y explicit
]
coordinates {
(1, 1.28)
(2, 1.28)
(3, 1.28)
(4, 1.29)
(5, 1.28)
(6, 1.28)
(7, 1.277)
(8, 1.277)
(9, 1.28)
(10, 1.29)
(20, 1.28)
(30, 1.28)
(40, 1.28)
(50, 1.28)
(60, 1.28)
(70, 1.29)
(80, 1.29)
(90, 1.29)
(100, 1.29)
(110, 1.29)
(120, 1.3)
(130, 1.3)
(140, 1.3)
(150, 1.3)
(160, 1.29)
(170, 1.3)
(180, 1.3)
(190, 1.3)
(200, 1.3)
};\addlegendentry{Log Reg val}
\addplot[
%dash pattern=on 4pt off 1pt on 4pt off 4pt,
%dashed,
color=yellow,
%mark=*,
very thick,
error bars,
y dir= both,
y explicit
]
coordinates {
(1, 1.28)
(2, 1.28)
(3, 1.28)
(4, 1.28)
(5, 1.28)
(6, 1.28)
(7, 1.28)
(8, 1.28)
(9, 1.28)
(10, 1.28)
(20, 1.28)
(30, 1.28)
(40, 1.28)
(50, 1.28)
(60, 1.28)
(70, 1.28)
(80, 1.28)
(90, 1.28)
(100, 1.28)
(110, 1.28)
(120, 1.28)
(130, 1.28)
(140, 1.28)
(150, 1.28)
(160, 1.28)
(170, 1.28)
(180, 1.28)
(190, 1.28)
(200, 1.28)
};\addlegendentry{NB val}
\addplot[
%dash pattern=on 4pt off 1pt on 4pt off 4pt,
%dashed,
color=green,
%mark=*,
very thick,
error bars,
y dir= both,
y explicit
]
coordinates {
(1, 2.34)
(2, 2.17)
(3, 2.13)
(4, 2.14)
(5, 2.14)
(6, 2.14)
(7, 2.14)
(8, 2.15)
(9, 2.15)
(10, 2.15)
(20, 2.14)
(30, 2.14)
(40, 2.17)
(50, 2.08)
(60, 2.08)
(70, 2.07)
(80, 2.06)
(90, 2.06)
(100, 2.06)
(110, 2.06)
(120, 2.05)
(130, 2.05)
(140, 2.05)
(150, 2.04)
(160, 2.04)
(170, 1.46)
(180, 2.04)
(190, 1.95)
(200, 1.49)
};\addlegendentry{Perc val}
\addplot[
%dash pattern=on 4pt off 1pt on 4pt off 4pt,
%dashed,
color=blue,
%mark=*,
very thick,
error bars,
y dir= both,
y explicit
]
coordinates {
(1, 1.28)
(2, 1.28)
(3, 1.28)
(4, 1.28)
(5, 1.28)
(6, 1.28)
(7, 1.28)
(8, 1.28)
(9, 1.28)
(10, 1.28)
(20, 1.28)
(30, 1.28)
(40, 1.28)
(50, 1.3)
(60, 1.327)
(70, 1.345)
(80, 1.3856)
(90, 1.378)
(100, 1.37)
(110, 1.367)
(120, 1.367)
(130, 1.36)
(140, 1.407)
(150, 1.42)
(160, 1.43)
(170, 1.44)
(180, 1.44)
(190, 1.44)
(200, 1.44)
};\addlegendentry{Lin SVC val}
\end{axis}
\end{tikzpicture}
}
\subfigure[]{
\begin{tikzpicture}[scale=0.8]
\begin{axis}[
legend style={legend pos=south east, font=\tiny},
xlabel={No. of Features},
ylabel={Accuracy},
xmin=-1, xmax=202,
ymin=0.2, ymax=0.5,
xtick={0, 20, 40,60,80,100,120,140,160,180,200},
ytick={0.0, 0.1, 0.2, 0.3, 0.4, 0.5, 0.6, 0.7, 0.8, 0.9, 1.0},
ymajorgrids=true,
grid style=dashed
]
\addplot[
%dash pattern=on 4pt off 1pt on 4pt off 4pt,
%dashed,
color=red,
%mark=*,
very thick,
error bars,
y dir= both,
y explicit
]
coordinates {
(1, 0.34)
(2, 0.34)
(3, 0.34)
(4, 0.35)
(5, 0.36)
(6, 0.36)
(7, 0.37)
(8, 0.38)
(9, 0.38)
(10, 0.39)
(20, 0.41)
(30, 0.42)
(40, 0.42)
(50, 0.42)
(60, 0.42)
(70, 0.42)
(80, 0.42)
(90, 0.42)
(100, 0.43)
(110, 0.43)
(120, 0.43)
(130, 0.43)
(140, 0.43)
(150, 0.43)
(160, 0.43)
(170, 0.43)
(180, 0.43)
(190, 0.43)
(200, 0.43)
};\addlegendentry{Log Reg val}
\addplot[
%dash pattern=on 4pt off 1pt on 4pt off 4pt,
%dashed,
color=yellow,
%mark=*,
very thick,
error bars,
y dir= both,
y explicit
]
coordinates {
(1, 0.34)
(2, 0.34)
(3, 0.34)
(4, 0.34)
(5, 0.34)
(6, 0.34)
(7, 0.34)
(8, 0.34)
(9, 0.34)
(10, 0.34)
(20, 0.34)
(30, 0.34)
(40, 0.34)
(50, 0.34)
(60, 0.34)
(70, 0.34)
(80, 0.34)
(90, 0.34)
(100,0.34)
(110, 0.34)
(120, 0.34)
(130, 0.34)
(140, 0.34)
(150, 0.34)
(160, 0.34)
(170, 0.34)
(180, 0.34)
(190, 0.34)
(200, 0.34)
};\addlegendentry{NB val}
\addplot[
%dash pattern=on 4pt off 1pt on 4pt off 4pt,
%dashed,
color=green,
%mark=*,
very thick,
error bars,
y dir= both,
y explicit
]
coordinates {
(1, 0.17)
(2, 0.24)
(3, 0.25)
(4, 0.25)
(5, 0.25)
(6, 0.25)
(7, 0.25)
(8, 0.25)
(9, 0.25)
(10, 0.25)
(20, 0.25)
(30, 0.25)
(40, 0.26)
(50, 0.26)
(60, 0.26)
(70, 0.26)
(80, 0.27)
(90, 0.27)
(100, 0.27)
(110, 0.27)
(120, 0.27)
(130, 0.27)
(140, 0.27)
(150, 0.27)
(160, 0.28)
(170, 0.34)
(180, 0.28)
(190, 0.30)
(200, 0.36)
};\addlegendentry{Perc val}
\addplot[
%dash pattern=on 4pt off 1pt on 4pt off 4pt,
%dashed,
color=blue,
%mark=*,
very thick,
error bars,
y dir= both,
y explicit
]
coordinates {
(1, 0.34)
(2, 0.34)
(3, 0.34)
(4, 0.34)
(5, 0.34)
(6, 0.34)
(7, 0.34)
(8, 0.34)
(9, 0.34)
(10, 0.34)
(20, 0.34)
(30, 0.34)
(40, 0.34)
(50, 0.34)
(60, 0.34)
(70, 0.34)
(80, 0.35)
(90, 0.35)
(100, 0.35)
(110, 0.35)
(120, 0.35)
(130, 0.35)
(140, 0.36)
(150, 0.36)
(160, 0.37)
(170, 0.37)
(180, 0.37)
(190, 0.38)
(200, 0.38)
};\addlegendentry{Lin SVC val}
\end{axis}
\end{tikzpicture}
}
\caption{Latent Semantic Indexing (LSI)}
\label{}
\end{figure}
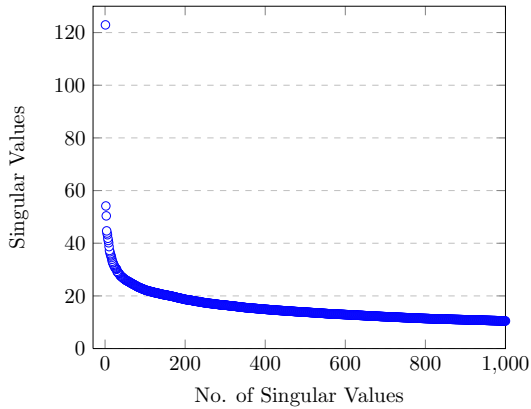
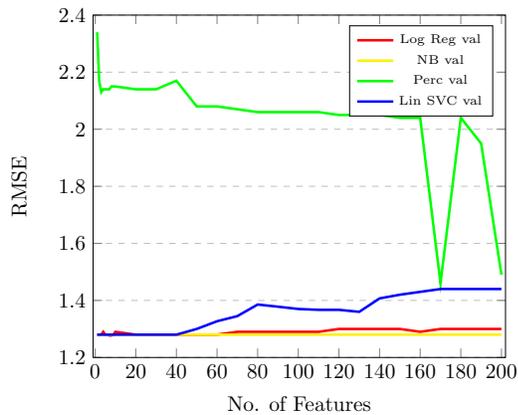
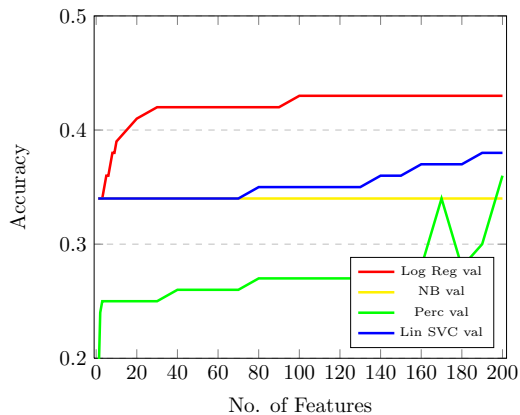

%%%%%%%%%%%%%%%%%%%%%%%%%%%%%%%%%%%%%%%%%%%%%%%%%%%%%%%%%%%%%%%%%%%%%%%%%%%%%%%%%%%%%%%%%%%%%%%%%%%%%
%%%%%%%%%%%%%%%%%%%%%%%%%%%%%%%%%%%%%%%%%%%%%%%%%%%%%%%%%%%%%%%%%%%%%%%%%%%%%%%%%%%%%%%%%%%%%%%%%%%%%
%%%%%%%%%%%%%%%%%%%%%%%%%%%%%%%%%%%%%%%%%%%%%%%%%%%%%%%%%%%%%%%%%%%%%%%%%%%%%%%%%%%%%%%%%%%%%%%%%%%%%

\subsection{The Best Model: Performance on the Test Set}
Based on the results in Figure 3, 4 and 5, we can see that Logistic Regression achieved the highest accuracy of 64\% using the top 10,000 Unigrams \& Bigrams as features, followed very closely by Linear SVC which achieved 63\% accuracy using the top 10,000 Unigrams \& Bigrams. 

Next, we evaluate these two systems on the test set. For Linear SVC, the RMSE and accuracy scores are 1.05 and 56\%, and for logistic regression, the scores are 0.92 and 54\%. These test-set scores are slightly worse than the validation-fold scores, and possibly indicate overfitting; to fix that, we would need to add/adjust the regularization parameters and re-run all the experiments. This is left as future work.

\section{Conclusions \& Future Work}

In this paper, we tackle the Review Rating Prediction problem for restaurant reviews on Yelp. We treat it as a 5-class classification problem, and examine various feature extraction and supervised learning methods to construct sixteen prediction systems. Experimentation and performance evaluation through $k$-fold cross validation yields one system, Logistic Regression on the set of top 10,000 features obtained from Unigrams \& Bigrams, that exhibits better predictive powers than the others. Our system can be used to generate star ratings on review websites where users can write free-form text reviews without giving a star rating. 

Though the methods tested in this paper are extensive, they are by no means exhaustive. In fact, there are many avenues for improvements and future work. We list them below.
\begin{enumerate}[itemsep=0mm]\vspace{-0.5cm}
\item We can try more sophisticated feature engineering methods, such as Parts-of-Speech (POS) tagging and spell-checkers, to obtain more useful $n$-grams. For example, instead of considering all possible bigrams, we can extract all the adjective-noun pairs or all the noun-noun pairs to get more meaningful 2-tuples. This would significantly help with the system's efficiency as well. Our current implementation in Python is quite slow  (each plot in Figures 3, 4 and 5 took 36 to 48 hours) because we deal with up to 32 million features. It is also memory-intensive. Choosing a smaller number of features more carefully will help tremendously with this bottleneck.
\item We can try more elaborate experiments for LSI that consider more than 200 features. Moreover, instead of performing singular value decomposition of unigrams only, we can add other $n$-grams. Another possibility is to perform SVD on specific text-constructs only, such as the nouns or the adjective-noun pairs. It would be interesting to see how the results change.
\item A possibly better alternative for logistic regression that should be tried is ordered/ordinal logistic regression \footnote{http://www.norusis.com/pdf/ASPC\_v13.pdf.}. It is an extension of logistic regression that specifically caters to classification problems where the class labels are ordered. So, in our case, this model takes into consideration the fact that the class labels 1 and 2 are closer to each other, than the labels 1 and 4. 
\item All the prediction models that we use in this paper are linear. That is, the hypothesis function is a linear function of the parameters. It would be interesting to experiment with non-linear models, e.g. polynomial regression, SVC with a non-linear kernel, etc. Alternatively, to get a non-linear decision boundary, we could find a linear decision boundary in an expanded feature space; for example, the feature vectors $\textbf{r}_i$ could be replaced with $\phi(\textbf{r}_i)$, where $\phi$ is called a \textit{feature mapping}. 
\item As mentioned in Section 5.5, our test-set performance was worse than the validation-fold performance, and this could be due to over-fitting. To fix this, we can introduce/adjust the regularization parameters in our models  using internal cross-validation  to get improved performance.
\item For feature extraction, we could try topic modelling techniques such as Latent Dirichlet Allocation \cite{blei03}, Non-negative Matrix Factorization \cite{tandon10} and/or Independent Component Analysis \cite{stone04}. We could also use one or more of the sentiment analysis techniques mentioned in Section 2,  and possibly combine them with the semantic analysis techniques we use in this paper to improve the results.
\item To get a more detailed performance evaluation, we can try other metrics, such as precision, recall, F-score and confusion matrix. Also, for probabilistic models, we can analyse the confidence scores.
\item We can compare our classifiers' preformance with the performance of the traditional recommendation techniques such as collaborative filtering.
\item We chose to do 3-fold cross validation in our experiments because 10-fold cross validation turned out to be very expensive in terms of time and memory. It would be worthwhile to try 10-fold cross validation and see if it yields different and/or better results. To deal with the time and RAM bottleneck, we could try parallel processing over clusters, such as those provided by Sharcnet\footnote{https://www.sharcnet.ca.}.  
\item Another avenue for future work is to test how our prediction models would perform on other business categories, such as shopping, travel, etc. 
\end{enumerate} 
%%%%%%%%%%%%%%%%%%%%%%%%%%%%%%%%%%%%%%%%%%%%%%%%%%%%%%%%%%%%%%%%%%%%%%%%%%%%%%%%%%%%%%%%%%%%%%%%%%%%%
%%%%%%%%%%%%%%%%%%%%%%%%%%%%%%%%%%%%%%%%%%%%%%%%%%%%%%%%%%%%%%%%%%%%%%%%%%%%%%%%%%%%%%%%%%%%%%%%%%%%%
%%%%%%%%%%%%%%%%%%%%%%%%%%%%%%%%%%%%%%%%%%%%%%%%%%%%%%%%%%%%%%%%%%%%%%%%%%%%%%%%%%%%%%%%%%%%%%%%%%%%%

%%%%%%%%%%%%%%%%%%%%%%%%%%%%%%%%%%%%%%%%%%%%%%%%%%%%%%%%%%%%%%%%%%%%%%%%%%%%%%%%%%%%%%%%%%%%%%%%%%%%%
%%%%%%%%%%%%%%%%%%%%%%%%%%%%%%%%%%%%%%%%%%%%%%%%%%%%%%%%%%%%%%%%%%%%%%%%%%%%%%%%%%%%%%%%%%%%%%%%%%%%%
%%%%%%%%%%%%%%%%%%%%%%%%%%%%%%%%%%%%%%%%%%%%%%%%%%%%%%%%%%%%%%%%%%%%%%%%%%%%%%%%%%%%%%%%%%%%%%%%%%%%%

%%%%%%%%%%%%%%%%%%%%%%%%%%%%%%%%%%%%%%%%%%%%%%%%%%%%%%%%%%%%%%%%%%%%%%%%%%%%%%%%%%%%%%%%%%%%%%%%%%%%%
%%%%%%%%%%%%%%%%%%%%%%%%%%%%%%%%%%%%%%%%%%%%%%%%%%%%%%%%%%%%%%%%%%%%%%%%%%%%%%%%%%%%%%%%%%%%%%%%%%%%%
%%%%%%%%%%%%%%%%%%%%%%%%%%%%%%%%%%%%%%%%%%%%%%%%%%%%%%%%%%%%%%%%%%%%%%%%%%%%%%%%%%%%%%%%%%%%%%%%%%%%%

\newpage
\bibliography{example_paper}
\bibliographystyle{plainnat}

\end{document}